\documentclass[sigconf]{acmart}

\usepackage[utf8]{inputenc}%(only for the pdftex engine)
\usepackage{caption}
\usepackage{subcaption}
\usepackage{makecell}
\usepackage{multirow}
\usepackage[linesnumbered,ruled,vlined]{algorithm2e}
\usepackage{amsmath}

\DeclareMathOperator{\argminA}{argmin}

\author{Yuting Zhan, Hamed Haddadi}
\affiliation{Imperial College London
  \country{UK}
}
\author{Afra Mashhadi}
\affiliation{University of Washington
  \country{USA}
}

\renewcommand\footnotetextcopyrightpermission[1]{} % removes footnote with conference info
\setcopyright{none}
\copyrightyear{2023}
\acmYear{2023}
\acmDOI{XXXXXXX.XXXXXXX}
\acmISBN{}
\acmConference[]{}{}{}
\begin{document}

\title{Privacy-Aware Adversarial Network in Human Mobility Prediction}

\begin{abstract}
As mobile devices and location-based services are increasingly developed in different smart city scenarios and applications, many unexpected privacy leakages have arisen due to geolocated data collection and sharing. 
%While these geolocated data could provide a rich understanding of human mobility patterns and address various societal research questions, privacy concerns for users' sensitive information have limited their utilization. 
User re-identification and other sensitive inferences are major privacy threats when geolocated data are shared with cloud-assisted applications.
Significantly, four spatio-temporal points are enough to uniquely identify 95\% of the individuals, which exacerbates personal information leakages.
To tackle malicious purposes such as user re-identification, we propose an LSTM-based adversarial mechanism with representation learning to attain a privacy-preserving feature representation of the original geolocated data (i.e., mobility data) for a sharing purpose.
These representations aim to maximally reduce the chance of user re-identification and full data reconstruction with a minimal utility budget (i.e., loss).
We train the mechanism by quantifying privacy-utility trade-off of mobility datasets in terms of trajectory reconstruction risk, user re-identification risk, and mobility predictability.
We report an exploratory analysis that enables the user to assess this trade-off with a specific loss function and its weight parameters.
The extensive comparison results on four representative mobility datasets demonstrate the superiority of our proposed architecture in mobility privacy protection and the efficiency of the proposed privacy-preserving features extractor.
We show that the privacy of mobility traces attains decent protection at the cost of marginal mobility utility.
Our results also show that by exploring the Pareto optimal setting, we can simultaneously increase both privacy (45\%) and utility (32\%).
\end{abstract}


\maketitle
%\pagestyle{plain}
%\section{Introduction}
\section{Introduction}

Geolocation and mobility data collected by location-based services (LBS)~\cite{huang2018location}, can reveal human mobility patterns and address various societal research questions~\cite{kolodziej2017local}. For example, Call Data Records (CDR) have been successfully used to provide real-time traffic anomaly and event detection~\cite{toch2019analyzing, wang2020deep}, and a variety of mobility datasets have been used in shaping policies for urban communities~\cite{ferreira_deep_2020} and epidemic management in the public health domain~\cite{oliver2015mobile,oliver2020mobile}. Moreover, users can benefit from personalized recommendations when they are encouraged to share their location data with third parties or other service providers (SPs, i.e., social platforms)~\cite{erdemir_privacy-aware_2020}. Human mobility prediction based on users' traces, a popular and emerging topic, supports a series of important applications ranging from individual-level recommendation systems to large-scale smart transportation~\cite{song2010limits}. For instance, one of the prerequisites for a successful LBS-recommendation system is the ability to predict users' activities or locations ahead of time, tracking their intentions and forecasting where they will go~\cite{gomes2013will}. 

While there is no doubt of the usefulness of predictive applications for mobility data, privacy concerns regarding the collection and share of individuals’ mobility traces have prevented the data from being utilized to their full potential~\cite{shokri2011quantifying, beresford2003location, krumm2009survey}. A mobility privacy study conducted by De Montjoye et al.~\cite{de2013unique} illustrates that four spatio-temporal points are enough to identify 95\% of the individuals, which exacerbates the user re-identification risk and could be the origin of many unexpected privacy leakages. 
Additionally, with increasingly intelligent devices and sensors being utilized to collect information about users' locations, a malicious third party can derive increasing intimate details about users' lives, from their social life to their preferences. 
Hence, a mechanism capable of decreasing the chance of user re-identification against malicious attackers or untrusted SPs can offer enhanced privacy protection in mobility data applications, as human mobility traces are highly unique.

In the past decade, the research community has extensively studied the privacy of geolocated data via various location privacy protection mechanisms (LPPM)~\cite{gedik2005location, gedik2007protecting}. Some traditional privacy-preserving approaches such as k-anonymity and geo-masking have shown to be insufficient to prevent users from being re-identified~\cite{de2013unique,song2010limits,gonzalez2008understanding, malekzadeh2020privacy}. Differential privacy (DP), another popular notion, is shown to be a limited metric for location trace privacy since temporal correlations are not taken into account~\cite{shokri2011quantifying}. ~\cite{erdemir_privacy-aware_2020} also states that DP and k-anonymity are meant to ensure the privacy of a single data point in time. In general, many DP for LBS (DP-L) mechanisms~\cite{andres2013geo, cunningham2021privacy, he2015dpt} attempt to protect the \emph{user's location} instead of protecting the \emph{user's identity}, which is outside the scope of our problem. However, we integrate one exemplary DP-L mechanism~\cite{andres2013geo} with a corresponding analysis of the user re-identification risk to further examine the effectiveness of the DP-L mechanism and achieve a more comprehensive comparison of our proposed mechanism.

\begin{figure}
    \centering
    \includegraphics[width=0.5\textwidth]{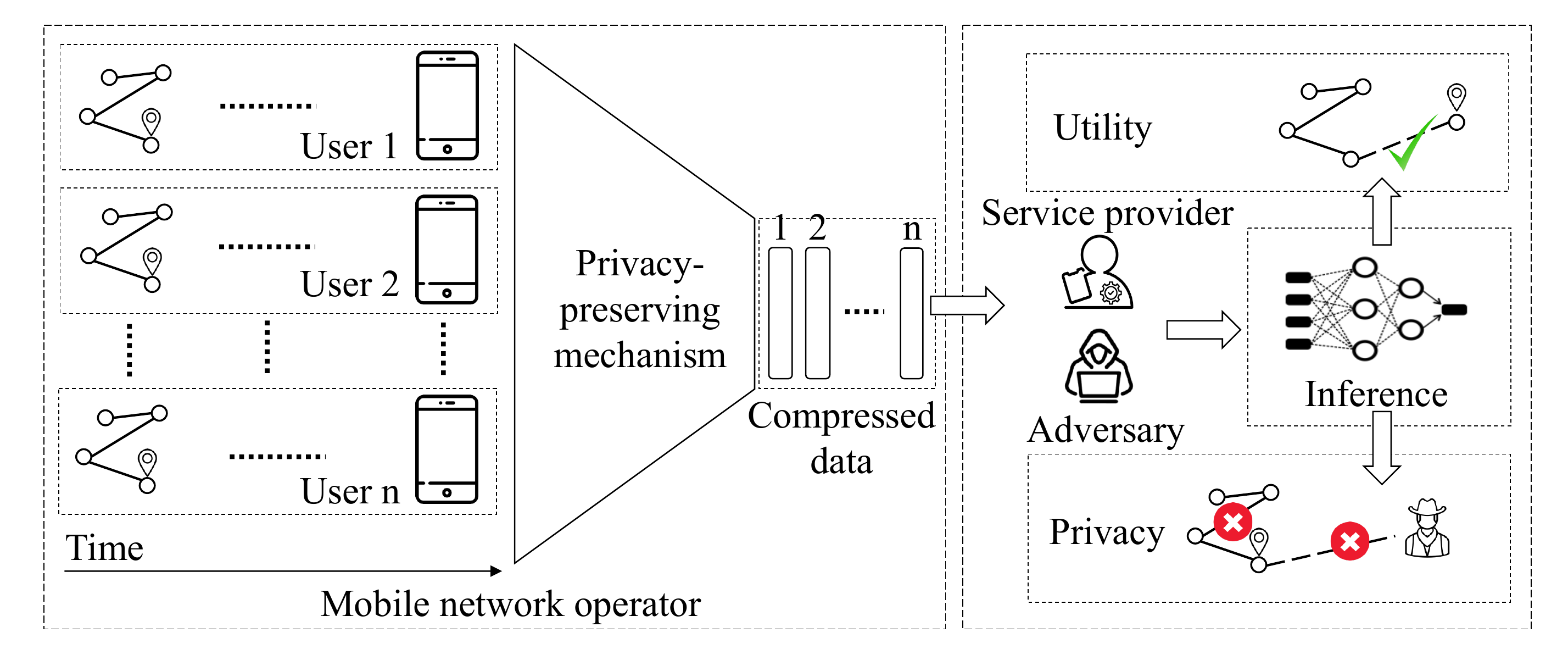}
    \caption{Privacy protection in users location data collection and sharing.}
    \label{fig:scenario}
\end{figure}

\begin{figure*}[t]
     \centering
     \begin{subfigure}[b]{0.72\textwidth}
         \centering
         \includegraphics[width=\textwidth]{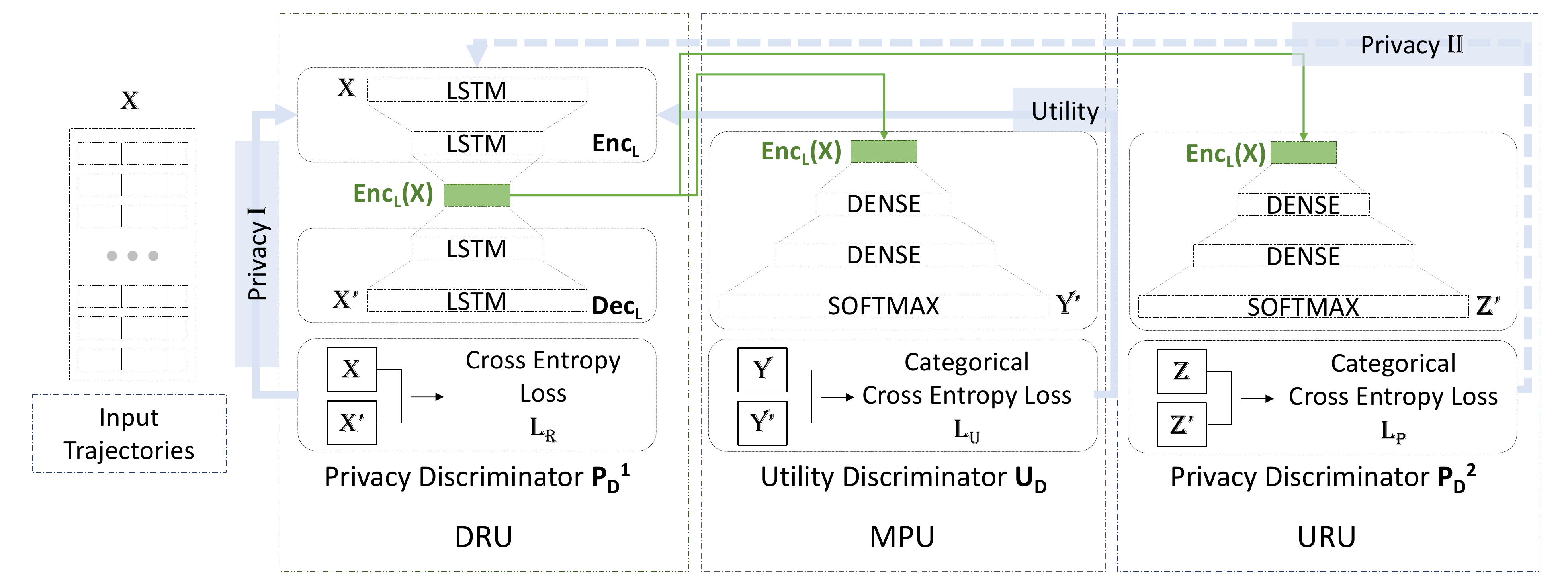}
         \caption{Mo-PAE}
         \label{fig:proposed}
     \end{subfigure}
     \begin{subfigure}[b]{0.21\textwidth}
         \centering
         \includegraphics[width=\textwidth]{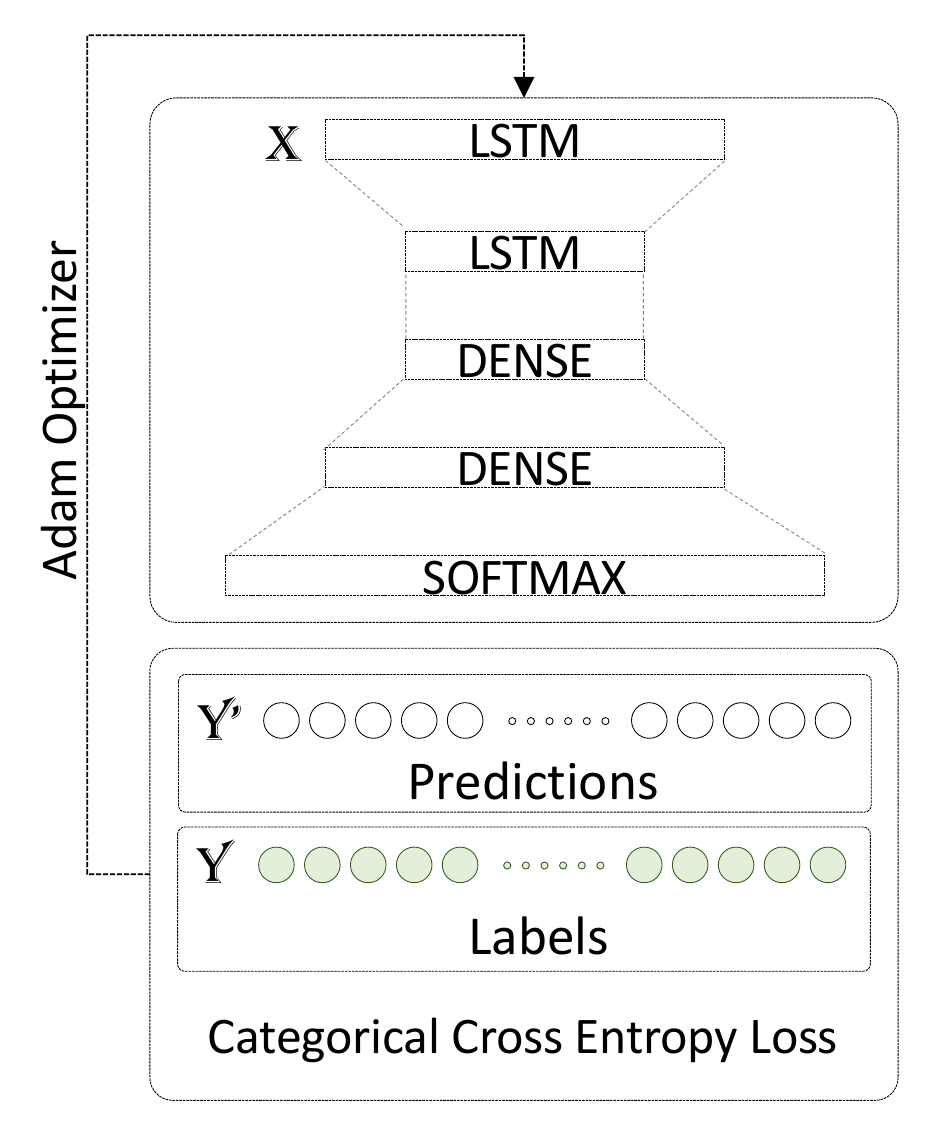}
         \caption{Optimal-IMs}
         \label{fig:baseline}
    \end{subfigure}
    \caption{(a) Schematic overview of the proposed privacy-preserving adversarial architecture (Mo-PAE), consisting of DRU, MPU, and URU. (b) The baseline LSTM network for optimal classifiers (Optimal-IMs).}
    %\Description{proposed architecture}
    \label{fig:architecture}
\end{figure*}

More recently, some related works have successfully applied machine-learning- or deep-learning-based approaches to explore effective LPPMs. Rao et al. proposed an model based on Generative Adversarial Network (GAN)~\cite{goodfellow2014generative} to generate privacy-preserving synthetic mobility datasets for data sharing and publication~\cite{rao2020lstm}. Feng et al. investigated human mobility data with privacy constraints via federated learning, achieving promising prediction performance while preserving the personal data on the local devices~\cite{feng2020pmf}. 
Though these works provide promising architectures to protect location privacy, the mobility data's privacy protection and utility degradation have not been fully investigated, especially in reducing the chance the user re-identification.
Our work extends these machine-learning-based mechanisms and explores the privacy-utility trade-off on mobility data in terms of declining the effectiveness of privacy inference attacks while maintaining its predictability.
Research on human mobility shows that the predictability of users' location trajectories or mobility, and the particular constraints of users' movements, are sufficient to reconstruct and/or identify anonymous or perturbed locations~\cite{shokri2012protecting}.
This specific confrontation makes the trade-off between mobility predictability and users' identity more challenging.
%Though these state-of-the-art models provide a reasonable balance between utility and privacy, the effectiveness of the privacy mechanism and utility metrics have not been fully investigated. 

%An appropriate and effective framework to allow researchers and practitioners to easily assess the trade-off between utility and privacy of mobility datasets at various granularity currently does not exist and can be highly impactful for the research community. 

Consider a scenario, shown in Figure~\ref{fig:scenario}, where users share their daily traces to a trusted mobile network operator, which then aggregates these traces in a privacy-preserving approach and sends them to third parties or other SPs with/without users' consent.
These users may want to minimize the risk of being re-identified and trajectory reconstructed by those who will access these released data. However, they would like to keep receiving potential effective services from SPs. Therefore, a privacy-preserving mechanism, which can release required information for the services (i.e., utility) while features or patterns that facilitate fully data reconstruction or user re-identification are obscured (i.e., privacy), is beneficial.
To this end, we propose a {\bf p}rivacy-aware {\bf a}dversarial network to train a feature extractor $Enc_L$ for {\bf mo}bility privacy, as shown in Figure~\ref{fig:architecture}, namely Mo-PAE. 

Mo-PAE is based on representation learning and aims to ease data sharing privacy concerns from privacy inference attacks.
Inspired by PAN~\cite{liu2019privacy} (privacy adversarial network), we employ adversarial learning to better balance the potential trade-off between privacy and utility. 
In contrast to PAN, which focuses on the privacy of images, our approach is designed for complex time-series data that exhibits spatial-temporal characteristics. 
At the core of our architecture lies an auto-encoder (AE) and long short-term memory (LSTM) layers with three branches, corresponding to the three training optimization objectives of the feature extractor $Enc_L$: i) to $maximize$ the loss associated with the reconstructed output by generative learning, ii) to $minimize$ the prediction loss using the learned representation from the $Enc_L$ by discriminative learning, and iii) to $minimize$ the percentage of users who are re-identifiable through their trajectories by discriminative learning. 
%We use Lagrange multipliers to vary the weights that are given to each of these objectives before combining them into a total loss, $L_{sum}$.
%The output of this model is a Pareto-Frontier analysis that would guide the user in investigating the trade-off between utility and privacy. 
We explore and quantify the privacy-utility trade-off achieve by Mo-PAE in terms of data reconstruction leakage (i.e., \textit{Privacy I}), user re-identification risk (i.e., \textit{Privacy II}), and mobility predictability (i.e., \textit{Utility}). 
%considering one the most popular human mobility scenarios (i.e., next location forecasting), 
%The benchmark comparisons are carried out with i). the DP mechanism developed on the idea from Geo-indistinguishability~\cite{andres2013geo} (namely GI-DP); ii). the machine-learning based mechanism based on GAN (namely TrajGAN~\cite{rao2020lstm}); iii). as well as the optimal inference models on raw data (namely Optimal-IMs). 
The results show that our proposed mechanism achieves a better privacy level with the same utility loss and vice versa.
%the synthetic dataset generated by LSTM-TrajGAN is not {\em pareto-optimal} in all four cases. That is, in the given spatial-temporal granularity, our proposed mechanism achieves a better privacy level for a dataset with the same utility value and vice versa.
The contributions of our work are the following:

\begin{itemize}

    \item We propose Mo-PAE to learn an effective privacy-preserving feature extraction \textbf{e}ncoder for mobility applications.
    
    \item We provide an extensive analysis of different mobility tasks and quantify the privacy and utility bound of the target mobility dataset, along with a trade-off analysis between these contrasting objectives. 
    
    \item We report the analysis of Mo-PAE by a comprehensive evaluation of four real-world representative mobility datasets.
    
    \item We compare our model with, i) a famous DP notion that developed on the idea from Geo-indistinguishability~\cite{andres2013geo} (namely GI-DP); 
    ii) a state-of-the-art GAN-based mechanism that attempts to generate synthetic privacy-preserving mobility data (namely TrajGAN~\cite{rao2020lstm}); iii) as well as the optimal LSTM-based inference model, and obtaining favorable results.
    
    %\item We offer the architecture as an open-source system to the researchers and practitioners.
    %\footnote{The open-source code of our framework will be publicly shared upon acceptance of this paper.}

\end{itemize}

The rest of this paper is structured as follows: 
we review the related work in Section \ref{RelW}; the proposed Mo-PAE is described in detail in Section \ref{DoF}; we describe the experimental settings in Section \ref{ExpS}; we demonstrate an evaluation of our mechanism over four mobility datasets with baseline comparisons in Section \ref{FraE}; Section \ref{sec:discussions} reports an in-depth discussion of our setting; finally, we conclude the paper with future work directions in Section \ref{Conl}.

\section{Related work}
\label{RelW}

\subsection{Notions of Location Privacy}

Diverse privacy notions, \emph{direct} or \emph{indirect}, for the LBSs have been proposed and evaluated in the literature. In~\cite{andres2013geo}, various \emph{direct notions} of location privacy and the techniques to achieve them are examined and concluded, including but not limited to expected distance error, $k$-anonymity, differential privacy (DP), and other location-privacy metrics.
First, the expectation of distance error reflects the accuracy when an adversary guesses the user's real location in a location-obfuscation mechanism by using the available side information. In~\cite{shokri2012protecting}, an optimal LPPM strategy and its corresponding optimal inference attack are obtained by formalizing the mutual optimization of user-adversary objectives (location privacy vs correctness of localization).
Second, $k$-anonymity is the most widely used privacy notion for the LBSs~\cite{wang2019achieving}. These systems aim to protect the \emph{user's identity}, requiring that the attacker cannot infer the correct user among a set of $k$ different users. 
Third, DP~\cite{dwork2008differential} is an emerging notion initially formulated in the context of statistical databases and aims to protect an individual's data while publishing aggregate information about the dataset. More precisely, a randomization mechanism $M$ gives $\epsilon$-differential privacy for all neighbouring datasets $D$ and $D'$, and the difference between $D$ and $D'$ is within a bound of $e^\epsilon$. One of the popular mechanisms to achieve DP perturb the original query result using random noise that is calibrated with the privacy budget $\epsilon$ and defines a global sensitivity for all neighbouring $D$ and $D'$~\cite{dwork2014algorithmic}. The work in~\cite{errounda2019analysis} reviews research works done in differential privacy targeted toward location data from a data flow perspective, including collection, aggregation, and mining.
~\cite{andres2013geo} proposed a Geo-indistinguishability notion based on differential privacy and a planar Laplace mechanism.
Significantly, different from the systems in $k$-anonymity category aim at protecting \emph{user's identity}, DP mechanisms are interested in protecting the \emph{user's locations}~\cite{andres2013geo, cunningham2021privacy, he2015dpt, gursoy2020utility}.
Apart from three mainstream approaches, the location cloaking mechanism tries to define the uncertainty region and measure the privacy by the size of the cloak and by the coverage of sensitive regions; the inaccuracy of the sensing technology tries to achieve a certain level of privacy by increasing uncertainty; and transformation-based approach tries to make user's location invisible to the service provider.

In the other hand, \emph{indirect} notions of location privacy arise with the emerging machine learning-based mechanism, which assesses the privacy guarantee by measuring the effectiveness of target inference attacks~\cite{chatzikokolakis2015geo, primault2014differentially, gursoy2018utility}. 
In general, for any LBS, their main privacy concerns can be concluded in two categories. One is the attack on the \emph{user's identity} which can be re-identified maliciously. For instance, even if the adversary is assumed to be unaware of the user identity of a trace, they can infer \emph{user's identity} or additional sensitive information due to the location information leakage based on publicly accessible background information.
The other attack is the one on \emph{user's location} while the adversary has legible access to \emph{user's identity}. In this manner, \emph{user's locations} are sensitive, which could exert a significant impact on other sensitive personal details, such as religious affiliation, sexual orientation, economic condition, health status, and so on.

In our work, we are interested in protecting \emph{user's identity} as the privacy scope, which is similar to the location privacy notion defined by the $k$-anonymity, and taking the real/distorted \emph{user's location} as input for the personal recommendation model to provide contextual services for their future travels. In general, DP paradigms have the most formal privacy guarantee than the others, however, they are not immune to inference attacks~\cite{clifton2013syntactic, hamm2017minimax}. We will also compare our proposed model with one popular DP paradigm on location privacy to illustrate the ineffectiveness of DP to our research question.
More details on our privacy definitions are in Section~\ref{DoF}.

\subsection{Location Privacy Preserving Mechanisms}
An effective location-privacy preserving mechanism (LPPM) must consider three fundamental elements: i). the privacy requirements of the users (namely \textit{privacy gain}); ii). the adversary's knowledge and capabilities; iii). and maximal tolerated service quality degradation stemming from the obfuscation of true locations (namely, \textit{utility loss})~\cite{shokri2012protecting}.
The literature on location privacy can be roughly classified into three categories: the design of LPPMs~\cite{bindschaedler2016synthesizing}; recovering actual user traces from anonymized or perturbed traces; the formal analysis and the search for an appropriate location privacy metric that allows for the fair comparison between LPPMs~\cite{shokri2012protecting, lippmann2000evaluating}.
Generally, typical LPPMs use some obfuscation methods - like spatial cloaking, cell, merging, location precision reduction, or dummy cells - to manipulate the probability distribution of user's location.
The most popular approach to protect location privacy is to send a space- or time-obfuscated version of the users' actual traces to the trusted or un-trusted third parties~\cite{shokri2012protecting}.
Xiao et al.~\cite{xiao2015protecting} investigate how to obtain location privacy under temporal correlations with an optimal DP-based LPPM.
Another mainstream approach tries to issue the dummy requests from fake locations to the services provider, the location privacy hence protected as these fake locations increase the uncertainty of the adversary about the users' real movements~\cite{shokri2012protecting}.
The other popular alternative utilizes mixed zones or silent periods to hide users' locations, as the adversary cannot link those who enter with those who exit the region when several users traverse the zone simultaneously.

In general, any LPPM would alter the location information, resulting in a severe distortion of data. Therefore, designing an optimized privacy-preserving algorithm with constrained utility degradation according to user privacy requirements is one critical dimension of LPPMs~\cite{bindschaedler2016synthesizing}.
There is no way to optimally address location privacy issues for all types of location-based systems, and the design of a specific LPPM requires carefully considering the application scenario and the realistic privacy requirements of mobile users~\cite{jiang2021location}. Hence, if a user prefers high service quality rather than the concerns of privacy leakage, then a more flexible system could be applied to guarantee the service quality. 
Our proposed model, in this way, can perform flexibly in application scenarios when users have different focuses on privacy or service.

\subsection{Privacy Preserving Techniques for Spatial-Temporal Data}
%To contextualize our work, we briefly review the current state of the art in machine learning-based privacy preservation techniques for mobility data.
%Current location privacy protection studies focus on two research streams.
Current privacy-preserving techniques for spatial-temporal data focus on two research streams.
One is the DP approach to grouping and mixing the trajectories from different users so that the identification of individual trajectory data is converted into a k-anonymity problem~\cite{aktay2020google,xiao2015protecting,andres2013geo}. 
For example, a recent Privacy-Preserving Trajectory Framework (PPTPF)~\cite{yang_pptpf_2021} applies the k-indistinguishability to anonymize trips for each user by condensing them into $k-1$ trajectories and determining $k-1$ anonymized clusters of trips. 
%This clustering approach condenses the sensitive location data of users to short yet meaningful trajectories, ensuring that each user’s trip maintains its original utility.

The other stream focuses on synthetic data generation~\cite{rezaei2018protecting,huang_variational_2019,choi2021trajgail, ijcai2018-530}. 
Synthetic data generation methods have been extensively studied in recent years as a way of tackling privacy concerns of location-based datasets. 
The majority of existing mobility synthesis schemes are mainly categorized into two approaches:
one is a more traditional, simulation-based approach, while the other is a more recent, neural network-based generative modeling approach that utilizes recurrent autoencoders and generative adversarial networks to produce realistic trajectories~\cite{shin2020user}.
Simulation-based approaches generate mobility traces by modeling overall user behavior as a stochastic process, such as a Markov chain model of transition probabilities between locations, and then drawing random walks,
potentially with additional stochastic noise added, as demonstrated in Xiao et al \cite{xiao_loclok_2017}.
These approaches require considerable feature engineering effort and struggle to capture longer-range temporal and spatial dependencies in the data~\cite{luca2021survey} and are thus limited in their ability to preserve
the utility of the original datasets.
In contrast, the generative neural network approach synthesizes user mobility traces by learning via gradient descent back-propagation, then the optimal weights are utilized for decoding a high-dimensional latent vector representation into sequences that closely resemble the original data. 
Such traces can maintain important statistical properties of the original data while taking advantage of noise introduced in the reconstruction process, to improve data subject anonymity.
Huang et al \cite{huang_variational_2019} demonstrates the use of a variational autoencoder network to reconstruct trajectory sequences,
while Ouyang et al \cite{ijcai2018-530} utilizes a convolutional GAN, but neither work directly makes a quantitative
assessment of the extent of privacy protection that their algorithms provide \cite{huang_variational_2019, ijcai2018-530}.
The TrajGAN by Rao et al \cite{rao2020lstm} is a state of the art example of the generative trajectory modeling approach, which quantifies its privacy protection by demonstrating a significant decline in the performance of a second user ID classifier model on the synthetic outputs compared to the original input trajectories.
For these reasons, we used TrajGAN as a baseline for comparison.

Our proposed model takes the neural network-based generative modeling approach, but differs from existing methods, where we utilize a combined, multi-task adversarial neural network to simultaneously reconstruct trajectories, predict next locations, and re-identify users, from the same learned latent vector representation. 
We seek an optimal trade-off between the three tasks' individual losses by optimizing a sum loss function with per-task weights, improving the controllability of the relative utility and privacy of the outputs.

%\subsection{Privacy Preserving Techniques for Spatial-Temporal Data}
%  A broader body of work has focused on privacy techniques for spatial-temporal data including but not limited to the CDR data.
%Prior to release, a dataset must balance between two conflicting objectives: Privacy and Utility, and requires some level of distortion to achieve their respective thresholds.
%A mainstream approach for implementing privacy is through Generalization \cite{nergiz_towards_nodate}. The technique is predicated on two inter-related mechanisms: clustering and alignment. Clustering aims at finding the best grouping of trajectories that minimized a predefined cost function, and the alignment process aligns trajectories in each group \cite{shaham_privacy_2019}. In 2009, $k$-Anonymity was adopted by a group of researchers \cite{nergiz_towards_nodate} for anonymization of spatio-temporal datasets. 
%In 2020,   an online private data release policy (PDRP) \cite{erdemir_privacy-aware_2020} that measures the privacy and utility trade-off was proposed. The process compares a distorted dataset with the original and minimizes the number of shared records while keeping the distortion below a certain threshold. The process begins with DBSCAN clustering and calculation of the Euclidean distance between the true location point and the closest centroid. Each time-series point from the result is considered as a state and placed into a Markov Chain to maintain time-series information. An additive multiplier is applied at each state to obfuscate the true position of the event.

\section{Design of the Architecture}
\label{DoF}

\subsection{Definition of Important Terms}

\subsubsection{Mobility Trace}

The raw geolocated data or other mobility data commonly contain three elements: user identifiers \textit{u}, timestamps \textit{t}, and location identifiers \textit{l}. Hence, each location record \textit{r} could be denoted as \emph{$r_i$} = [\textit{$u_i$}, \textit{$t_i$}, \textit{$l_i$}], while each location sequence \emph{S} is a set of ordered location records \emph{$S_n$} =~\{\textit{$r_1$, $r_2$, $r_3$, $\cdots$, $r_n$}\}, namely \textit{mobility trace}. In this paper, we assume that different users' mobility traces are collected and aggregated by trusted telecom operators or social platforms and shared with third-party SPs (trusted or untrusted). Therefore, given the past mobility trace \emph{$S_n$} =~\{\textit{$r_1$, $r_2$, $r_3$, $\cdots$, $r_n$}\}, the mobility prediction task is to infer the most likely location \emph{l$_{n+1}$} at the next timestamp \emph{t$_{n+1}$}. The data fed into the proposed architecture are a list of traces with specific sequence length (\emph{SL}), that is \{$S_{sl}^1$, $S_{sl}^2$, $S_{sl}^3$, $\cdots$, $S_{sl}^j$\}. For instance, if the sequence length is 10, that indicates each trace contains 10 history location records \textit{r}, \emph{$S_{10}$} = \{$r_1$, $r_2$, $r_3$, $\cdots$, $r_{10}$\}, and $SL=10$.

\subsubsection{User Re-identification}

The user re-identification risk arises because of the high uniqueness of human traces~\cite{de2013unique} and could be the origin of many unexpected privacy leakages. We assume each trace \emph{S} is originally labeled with a corresponding user identifier \textit{u}, and the user re-identification is to infer the user \textit{u} to whom the target trace \emph{$S_n$} = \{$r_1$, $r_2$, $r_3$, $\cdots$, $r_n$\} belongs. We thereby leverage the user identifiers \textit{u} as the ground-truth values for the user identity classes. This identity information is what we want to protect in the proposed adversarial network; that is, the entire network should convey as little user identifiable information as possible to decrease the user re-identification accuracy. At the same time, the built-in adversary tries to achieve maximum accuracy.

\subsection{Problem Definitions}
Research on human mobility or LBS shows that the predictability of users' traces or mobility, and the particular constraints of users' movements, are sufficient to reconstruct and/or identify anonymous or perturbed locations~\cite{shokri2012protecting}. This confrontation makes the trade-off between keeping mobility predictability and reducing the chance of user re-identification more interesting. For instance, an adversary can re-identify anonymous users' traces given the users' mobility profile~\cite{de2019give}; infer the users' next activities from the frequency of location visits~\cite{gomes2013will}; even obtain the personal home or working address from the trajectories~\cite{primault2018long}. A number of works on location privacy protection try to evaluate various questions about location information leakage~\cite{shokri2012protecting}. In this work, we design a model to protect location privacy regarding users' identity and data integrity while simultaneously minimizing the service quality (i.e., accuracy of next location forecasting) degradation stemming from the obfuscation of true data. Before describing our proposed model in detail, we first give a brief problem definition of the trade-offs of mobility data between utility and privacy.

\textbf{Data Utility:}
Mobility datasets are of great value for understanding human behavior patterns, smart transportation, urban planning, public health issue, pandemic management, etc. Many of these applications rely on the next location forecasting of individuals, which in the broader context, can provide an accurate portrayal of citizens' mobility over time. Mobility prediction not only can be analyzed to understand personalized mobility patterns but can also inform the allocation of public resources and community services. We focus on the capability of \emph{mobility prediction} (\emph{next location forecasting}) in this paper, and leverage the accuracy of the prediction as an important metric for quantifying the data utility. Hence, we define \textit{Utility} as follows:

\textit{Utility} (\textit{U}): the mobility predictability (i.e., forecasting accuracy). A higher forecasting accuracy indicates higher utility. \textit{Utility loss} then represents the accuracy degradation when the proposed privacy protection mechanism is applied.

\textbf{Privacy Protection:}
Mobility datasets shared with trusted or un-trusted third parties/SPs can be utilized to attain personalized preferences. However, these shared traces also contain sensitive information. With increasing intelligent devices and sensors being utilized to collect information about human activities, the traces also increasingly expose intimate details about users' lives, from their social life to their preferences. The capability of user re-identification is important to balance the risks and benefits of mobility data usage, for all data owners, third parties, and researchers. In our proposed architecture, we design two built-in adversaries to infer the ability of generated features to protect users' sensitive information. We then leverage the reduction of data reconstruction risk and user re-identification risk as our privacy metrics to evaluate our proposed architecture. Hence, the definition of \textit{Privacy} is summarized as follows:

\textit{Privacy I} (\textit{PI}): the distance between the reconstructed data $X'$ and the original data $X$, that is, information loss in the reconstruction process. More information loss indicates higher privacy protection.

\textit{Privacy II} (\textit{PII}): the decline of user re-identification accuracy, that is, the user de-identification effectiveness. The higher degradation of user re-identification accuracy indicates higher privacy protection, as fewer users would be re-identified by the adversary during the privacy attack.

Similar to the definition of the \textit{utility loss}, \textit{privacy gain} (in terms of \textit{PI} and \textit{PII}) quantifies the privacy information protected by the designed privacy-preserving mechanism.

\textbf{Privacy vs. Utility Trade-off:}
There is an inherent trade-off between location privacy protection and utility degradation~\cite{jiang2021location}. That is, achieving a better level of privacy protection may require sacrificing the service quality provided by the data. Such trade-off is omnipresent in various privacy protection mechanisms, especially in location obfuscation mechanisms. Higher privacy protection is achieved when the probability of an adversary inferring the true location of the user decrease, however, the result of a query based on the obfuscated location is significantly different from the actual interest of the user. The privacy-utility trade-off, hence, need to be examined and analyzed to guarantee the efficiency of the privacy protection mechanism. 

To be specific, a more obfuscated dataset will tend to perform better at preserving privacy, but worse at preserving utility, and vice versa. Hence, monitoring these two performance metrics in tandem allows users to select the optimal privacy-utility trade-off for their use cases, given their hyperparameter selections. Our Mo-PAE model is designed to train a features Encoder $Enc_L(X)$ that could convey more information on the utility but less on privacy and investigate a better trade-off between them. More details will be discussed in the following sub-section.

\subsection{Architecture Overview}
\label{section: module overview}

We discuss the overall design of the Mo-PAE before detailly reviewing the functionality of each unit.

\subsubsection{Overall Design}
\hfill

Our proposed \textbf{p}rivacy-preserving \textbf{a}dversarial feature \textbf{e}ncoder on mobility data, denoted as the \emph{Mo-PAE}, is based on representation learning and adversarial learning and aims to ease data sharing privacy concerns. Figure~\ref{fig:architecture} presents the basic workflow of the proposed Mo-PAE. It composes of three crucial units: data reconstruction risk unit (DRU), mobility prediction unit (MPU), and user re-identification risk unit (URU). 

When three units train concurrently, the MPU is regarded as the \textit{utility discriminator} $U_D$, while DRU and MPU act as two built-in adversaries and are regarded as the two \textit{privacy discriminators}, $P_D^1$ and $P_D^2$, respectively. The built-in adversary has been used as an effective adversarial regularization to prevent inference attacks, \textit{e.g.} in the classification setting~\cite{nasr2018machine} or in various GAN models~\cite{mukherjee2021privgan} for privacy-preserving purpose. We design two built-in adversaries to infer the ability of generated features to protect user's sensitive information, and they are trying to maximize the accuracy of privacy inference tasks in training. At the same time, an encoder $Enc_L$ is trained to produce feature representations $f$ from the target mobility data by jointly optimizing these extracted feature weights using the combined losses of the DRU, MPU, and URU simultaneously, during adversarial training. Therefore, in the Mo-PAE, the encoder $Enc_L$, and three discriminators $U_D$, $P_D^1$, $P_D^2$ play a multi-player game to minimaximize the value function $V(Enc_L, U_D, P_D^1, P_D^2)$:

\begin{equation}
\begin{aligned}
     & \underset{Enc_L, U_D}{\min} \underset{P_D^1, P_D^2}{\max} V(Enc_L, U_D, P_D^1, P_D^2) = \mathbb{E}_{x\sim X} [log U_D(Enc_L(x))] +\\ &  \mathbb{E}_{x\sim X} [log(1- P_D^1(Enc_L(x)))] + 
      \mathbb{E}_{x\sim X} [log(1- P_D^2(Enc_L(x)))]
\end{aligned}
\label{equ:adv}
\end{equation}

%The encoder $Enc_L$ is trained to achieve a better trade-off between service quality and user privacy budgets by extracting more information about mobility predictability but less about user privacy.
As described in the Eq.\ref{equ:adv}, we design a multi-task adversarial network to learn an LSTM-based encoder $Enc_L(X;\theta)$ with parameter set $\theta \in \Theta$, 
which can generate the optimized feature representations $f=Enc_L(X;\theta)$ via lowering the privacy disclosure risk of user identification information and improving the task accuracy (i.e., mobility predictability) concurrently.
Two potential malicious privacy leakages from URU and DRU, are attempted to retrieve sensitive information from the feature representations $f$. As built-in adversaries, they have full access to the feature representations $f$ and the entire encoder network with parameter set $f=Enc_L(X;\theta)$. In this manner, they have the optimal decoder setting.
Hence, the notion of privacy (\emph{privacy gain}), is measured by the decline of the effectiveness of target inference attacks (i.e., user re-identification attack and data reconstruction attack).

\begin{algorithm}[t]

\SetKwInOut{Input}{Input}\SetKwInOut{Output}{Output}
\SetAlgoLined
	\Input{Mobility data \textbf{X}, real mobility prediction labels \textbf{Y}, real user identification labels \textbf{Z}, weights: $\lambda_1$, $\lambda_2$, $\lambda_3$} 
	\Output{Adversarial Encoder $Enc_L(X; \theta_E, \theta_R, \theta_U, \theta_P)$}
	Initialize model parameters $\theta_E, \theta_R, \theta_U, \theta_P$;\\
	 \For{n epochs}{ 
	 	\For{$k = 1$, $\cdots$$, K_t$}{
	 	 1. Sample a mini-batch of mobility trajectories x, prediction labels y, identification labels z\\
	 	 2. Update $\theta_E$ with Adam optimizer on mini-batch loss $L_{sum}(\theta_E, \theta_R, \theta_U, \theta_P, \lambda_1, \lambda_2, \lambda_3)$\\
	 	 3. Update $\theta_R$ with Adam optimizer on mini-batch loss $L_R(f; \theta_R)_{(x, \hat{x})}$: $\min L_R$\\
	 	 4. Update $\theta_U$ with Adam optimizer on mini-batch loss $L_U(f; \theta_U)_{(y, \hat{y})}$: $\min L_U$\\
	 	 5. Update $\theta_P$ with Adam optimizer on mini-batch loss $L_P(f; \theta_P)_{(z, \hat{z})}$: $\min L_P$\\
  	    }
  	    Update with the gradient descent on $L_{sum}(\theta_E, \theta_R, \theta_U, \theta_P, \lambda_1, \lambda_2, \lambda_3)$: $\min L_{sum}$
 	 } 
    \caption{Training of the Mo-PAE (\emph{Model II})}
    \label{algo_disjdecomp} 
\end{algorithm}

\subsubsection{Details of Mo-PAE}
\hfill

We define the raw mobility data we want to protect as $\mathcal{X}$, trained features as $\mathcal{F}$, and reconstructed data as $\mathcal{X'}$. Given mobility raw data $\mathcal{X}$ for $P_D^1$ (DRU), the ground-truth label $z_i$ for $P_D^2$ (URU), and the ground-truth label $y_i$ for utility $U_D$ (MPU), we train the encoder $Enc_L$ to learn the representation $\mathcal{F}=Enc_L(\mathcal{X}; \theta_E)$. We design a specific loss function, namely \textit{sum loss}  $\mathcal{L}_{sum}$, for this optimization process.

Specifically, when reconstructing the data $\mathcal{X'}$, a decoder $Dec_L$ attempts to recreate the data based on the features $\mathcal{F}$, that is $Dec_L(\mathcal{F}; \theta'_D): \mathcal{F}\rightarrow \mathcal{X'}$. This DRU, the first privacy discriminator $P_D^1$, is trained as a built-in adversary and tries to achieve sensitive information as much as possible. Hence, the DRU is primarily trained by minimizing the reconstruction loss $\mathcal{L_R}$: 
\begin{equation}
 \min \mathcal{L_R} \Rightarrow  \mathcal{L_R} = d(\mathcal{X}, \mathcal{X'}) = \underset{\mathcal{F}; \theta'_R}{\argminA} \| Dec_L(\mathcal{F}, \theta'_R) - \mathcal{X}\|^2
\end{equation}

The URU, the second privacy discriminator $P_D^2(\mathcal{F}; \theta')$, is trained to re-identify whom the target trajectory belongs to. 
It outputs a probability distribution of predicted user identifiers among Z potential classes.
Then in this privacy discriminator, the user re-identification loss $\mathcal{L_P}$ is primarily trained to minimize, denoted as $\min\mathcal{L_P}$:
\begin{equation}
\min \mathcal{L_P} \Rightarrow  \mathcal{L_P} = \underset{\mathcal{F}; \theta'_P}{\argminA} \sum_{i=1}^{\mathbf{Z}} z_i\text{log}(P_D^1(\mathcal{F}; \theta'_P))
\end{equation}

The MPU, the utility discriminator $U_D(\mathcal{F}; \theta')$, is trained to output a probability distribution of the next location of interest, and this distribution has Y potential classes.
Discriminative training of $U_D$ maximizes the prediction accuracy by minimizing the utility loss $\mathcal{L_U}$ concurrently with minimizing the $\mathcal{L}_{sum}$, denoted as $\min\mathcal{L_U}$. 
\begin{equation}
\min\mathcal{L_U} \Rightarrow \mathcal{L_U} =  \underset{\mathcal{F}; \theta'_U}{\argminA} \sum_{i=1}^{\mathbf{Y}} y_i\text{log}(U_D(\mathcal{F}; \theta'_U))
\end{equation}

The overall training is to achieve a privacy-utility trade-off by adversarial learning on $\mathcal{L_R}$, $\mathcal{L_U}$, and $\mathcal{L_P}$, concurrently. The encoder $Enc_L(\mathcal{X};\theta_E)$ should satisfy high predictability (\textit{min}~$\mathcal{L_U}$) and low user re-identification accuracy (\textit{max}~$\mathcal{L_P}$) of the mobility data when maximizing the reconstruction loss (\textit{max}~$\mathcal{L_R}$) in reverse engineering, where the training objective transformed from Eq.\ref{equ:adv} can be written as:
\begin{equation}
    \min \mathcal{L}_{sum} = \underset{\mathcal{L}_{U}}{\text{min}}\ \underset{\mathcal{L}_{R}, \mathcal{L}_{P}}{\text{max}} (\sum_{x=i}^{\mathcal{X}} \left(\mathcal{L}_{U} \left(f_i\right), \mathcal{L}_{P}\left(f_i\right), \mathcal{L}_{R}\left(f_i\right)\right))
\label{equ:sumloss1}
\end{equation}

We use Eq.~\ref{equ:sumloss1} to guide the first version of Mo-PAE, denoted as \emph{Model I}.
In order to fully investigate the range of trade-offs, we leveraged the Lagrange multipliers~\cite{beavis1990optimisation} as hyperparameters to control the privacy-utility tradeoffs in the Mo-PAE, and this weighted-controlled model is denoted as \emph{Model II}. Accordingly, the optimization function of the training objective is:
\begin{equation}
\begin{aligned}
    & \min \mathcal{L}_{sum}  = \underset{\mathcal{L}_{U}}{\text{min}}\ \underset{\mathcal{L}_{R}, \mathcal{L}_{P}}{\text{max}} (\sum_{x=i}^{\mathcal{X}} \left(\lambda_1\mathcal{L}_{R} \left(f_i\right), \lambda_2\mathcal{L}_{U}\left(f_i\right), \lambda_3\mathcal{L}_{P}\left(f_i\right)\right)) \\
    &= \underbrace{- \lambda_1 \left(\max\mathcal{L_R}\left(f_i\right)\right)}_{\text{Privacy\ I}} + \underbrace{\lambda_2 \left(\min\mathcal{L_U}\left(f_i\right)\right)}_{\text{Utility}} 
    \underbrace{- \lambda_3 \left(\max\mathcal{L_P}\left(f_i\right)\right)}_{\text{Privacy\ II}} \\ 
    &= -\lambda_1  \| Dec_L(\mathcal{F}) - \mathcal{X}\|^2 + \lambda_2  (\sum_{i=1}^{\mathbf{Y}} y_i\text{log}(U_D(\mathcal{F}))) \\
    &- \lambda_3  (\sum_{i=1}^{\mathbf{Z}} z_i\text{log}(P_D(\mathcal{F})))
\end{aligned}
\label{equ:sumloss}
\end{equation}
where $y_i$ is the ground-truth label for \textit{Utility}, $z_i$ is the ground-truth value for \textit{Privacy II}; $\lambda_1$, $\lambda_2$ and $\lambda_3$ are non-negative, real-valued weights, as the hyperparameters that control the privacy-utility trade-off in the Mo-PAE.

As shown in the Algorithm~\ref{algo_disjdecomp}, the gradient of the loss (i.e., $\theta_E$, $\theta_R$, $\theta_U$, $\theta_P$) back-propagates through the LSTM network to guide the training of the encoder $Enc_L$. The encoder is updated with the \textit{sum loss} function $\mathcal{L}_{sum}$ until convergence. It is tricky to practically investigate all possible weight combinations, we look for the optimal combinations through training~\cite{malekzadeh2020privacy} with the Eq.\ref{equ:sumloss} by brute-force evaluation. Then we approximate the required data utility reserved and reformulate the optimization problem in Eq.\ref{equ:sumloss} as a maxima privacy optimization problem.
\begin{equation}
\begin{aligned}
     & \underset{Enc_L}{\min} \underset{P_D^1, P_D^2}{\max} V_{\lambda \rightarrow U_D}(Enc_L, P_D^1, P_D^2)  = \mathbb{E}_{x\sim X} [log(1- P_D^1(Enc_L(x)))] \\ &  + \mathbb{E}_{x\sim X} [log(1- P_D^2(Enc_L(x)))]
\end{aligned}
\label{equ:adv2}
\end{equation}
Additionally, another key contribution is the flexibility of the \textit{sum loss} function $\mathcal{L}_{sum}$ which could be regulated to satisfy different requirements on privacy protection level and service quality. That is, different combinations of weights control the relative importance of each unit and guide the overall model to find the maxima or minima given the specific trade-off choices.

\begin{table*}[t!]
    \centering
    \begin{tabular}{|c|cccc|cc|cc|}
     \hline
\multirow{2}{*}{\shortstack{Dataset-City}} & \multicolumn{4}{c|}{Bounding Box} & \multicolumn{2}{c|}{Record Counts} & \multicolumn{2}{c|}{Number}\\
\cline{2-9}
         & \multicolumn{2}{c}{Latitude} & \multicolumn{2}{c|}{Longitude} & Train  & Test & User ID & POI \\
         \hline
        MDC-Lausanne & 46.50 & 46.61 & 6.58 & 6.73 & 77393 & 19429 & 143 & 149 \\
        %\hline
        Priva'Mov-Lyon & 45.70 & 45.81 & 4.77 & 4.90 & 62077 & 16859 & 58 & 129 \\
        %\hline
        GeoLife-Beijing & 39.74 & 40.07 & 116.23 & 116.56 & 95038 & 24578 & 145 & 960 \\
        %\hline
        FourSquare-NYC & 40.55 &  40.99 & -74.28 & -73.68 & 43493 & 11017 & 466 & 1712 \\
        \hline
    \end{tabular}
    \caption{Overview of four mobility datasets after pre-processing. The bounding box represents the range of the considered locations/traces.}
    \label{tab:datasets}
\end{table*}

\subsubsection{Composition Units of Architecture}
\hfill

\textit{I. Mobility Prediction Unit (MPU)}:  

The MPU unit is composed of three parts, the input part with the multi-modal embedding of trace information, the sequential part with LSTM layers~\cite{hochreiter1997long}, and an output part with the softmax activation function. As per the definition mentioned earlier, the traces in this work are shown as location sequences \emph{S}. First, the location identifiers \textit{l} and timestamps \textit{t} are converted into one-hot vectors. We then employ LSTM layers to model the mobility patterns and sequential transition relations in these mobility traces. As a prominent variant of the recurrent neural network, LSTM networks exhibit brilliant performance in modeling the entire data sequences, especially for learning long-term dependencies via gradient descent~\cite{zhan2019towards}. Following the sequential module, the softmax layer outputs the probability distribution of the prediction results. This probability distribution is converted to the top-n accuracy metrics to illustrate the unit performance.

\textit{II. Data Reconstruction Risk Unit (DRU)}:

The DRU is the encoder $Enc_L$ unit in reverse, also denoted as $Dec_L$, which is regarded as the first \textit{privacy discriminator} $P_D^1$ in the proposed architecture. 
It is designed to evaluate the distance $d(\cdot,\cdot)$ (i.e., \textit{Privacy I}) between the reconstructed data $X'$ and the original input data $X$. A malicious party is free to explore any machine learning model and reconstruct the data if they have the shared extracted features $f$. We use a layer-to-layer reverse architecture of our encoder $Enc_L$ to build the \textit{data reconstruction unit} to act as a robust built-in adversary. To compare with baseline models and keep the comparison in a line, we measure the distance $d(\cdot,\cdot)$ between the $X$ and $X'$ by leveraging the \textit{Euclidean} and \textit{Manhattan} distance as our metrics. Both of them are widely used in location privacy literature~\cite{andres2013geo, erdemir_privacy-aware_2020}. 
%For instance, the work introducing Geo-Indistinguishability~\cite{andres2013geo} utilizes a privacy level that depends on the \textit{Euclidean} distance.

\textit{III. User Re-identification Risk Unit (URU)}: 

The URU is regarded as the second \textit{privacy discriminator} $P_D^2$ in the proposed architecture. 
The unit is composed of three parts, the input part with the one-hot embedding of user identity, the sequential part with LSTM layers, and an output part with softmax function. First, the user identity list is converted into one-hot vectors. Similar to the MPU, the URU also applies LSTM layers to better extract the spatial and temporal characteristics of the context. A softmax function with a cross-categorical entropy loss function is applied to output a categorical probability distribution of the user re-identification task. We then use the top-N accuracy of this classifier as the metrics of user re-identification privacy risk (i.e., \textit{Privacy II}). The more accurately a classifier can re-identify the user when given a trajectory, the higher the risk of disclosing private data.
Same as $P_D^1$, $P_D^2$ are designed as the built-in adversary to infer the ability of generated features in protecting users' sensitive information.

The overall architecture eventually learns to fool both built-in adversaries, $P_D^1$ and $P_D^2$, while maintaining the mobility predictability. In this manner, both adversaries are assumed to be free to access the exclusive feature representations and the entire encoder network, which allows them to have the optimal decoder setting. We will discuss the effectiveness of two privacy inference attacks in Section~\ref{UPA}.

\section{Experimental Setting}
\label{ExpS}

\subsection{Datasets}
Experiments are conducted on four representative mobility datasets: Mobile Data Challenge Dataset (MDC)~\cite{laurila2012mobile}, Priva'Mov~\cite{mokhtar2017priva}, GeoLife~\cite{zheng2011geolife}, and FourSquare~\cite{dingqi_yang_modeling_2015}. Once imported into our architecture, each dataset was filtered and preprocessed individually to derive their respective train and test sets illustrated in Table~\ref{tab:datasets}. Each bounding box defines the grid size and the grid granularity is 0.01 degrees.

\noindent 
\textbf{MDC}: it is recorded from 2009 to 2011 and contains a large amount of continuous mobility data for 184 volunteers with smartphones running a data collection software, in the Lausanne/Geneva area. 
Each record of the \textit{gps-wlan} dataset represents a phone call or an observation of a WLAN access point collected during the campaign~\cite{laurila2012mobile}.  

\noindent 
\textbf{Priva'Mov}: the PRIVA'MOV crowd-sensing campaign took place in the city of Lyon/France from October 2014 to January 2016. Data collection was contributed by roughly 100 participants including university students, staff, and family members. The crowd-sensing application collected GPS, WiFi, GSM, battery, and accelerometer sensor data. For this paper, we only used the GPS traces from the dataset ~\cite{mokhtar2017priva}. 

\noindent
\textbf{GeoLife}: it is collected by Microsoft Research Asia from 182 users in the four and a half year period from April 2007 to October 2011 and contains 17,621 trajectories \cite{zheng2011geolife}. 
This dataset recorded a broad range of users’ outdoor movements, including life routines like going home and going to work and some entertainment and sports activities, such as shopping, sightseeing, dining, hiking, and cycling. It is widely used in many research fields, such as mobility pattern mining, user activity recognition, location-based social networks, location privacy, and location recommendation.

\noindent
\textbf{FourSquare NYC}: it contains check-ins in NYC and Tokyo collected during the approximately ten months from 12 April 2012 to 16 February 2013, containing 227,428 check-ins from 1,083 subjects in New York City \cite{dingqi_yang_modeling_2015}.

\subsection{Baseline Models}
\label{sec:baseline}

\subsubsection{I. Optimal Inference Models (Optimal-IMs)} 
Optimal-IMs comprise three independent inference models: data reconstruction model, mobility prediction model, and user re-identification model. Each model has a similar layer design as the corresponding unit in the Mo-PAE, however, these three models are completely independent and have no effect on each other. Unlike the Mo-PAE, which leverages adversarial learning to finally attain an extracted feature representation $f$ that satisfies the utility requirements and privacy budgets simultaneously, the Optimal-IMs are only trained for optimal inference accuracy at the individual tasks to characterize the original data.

\subsubsection{II. LSTM-TrajGAN (TrajGAN)~\cite{rao2020lstm}} It is an end-to-end deep learning model to generate synthetic data which preserves essential spatial, temporal, and thematic characteristics of the real trajectory data. Compared with other standard geomasking methods, TrajGAN can better prevent users from being re-identified. The TrajGAN work claims to preserve essential spatial and temporal characteristics of the original data, verified through statistical analysis of the generated synthetic data distributions, which aligns with the mobility prediction-based utility metric in our work.
Hence, we train an optimal mobility prediction model for each dataset and evaluate the mobility predictability of synthetic data generated by the TrajGAN. In contrast to the TrajGAN that aims to generate synthetic data, our proposed Mo-PAE is training an encoder $Enc_L$ that forces the extracted representations \textit{f} to convey maximal utility while minimizing private information about user identity via adversarial learning.

\subsubsection{III. GI-DP~\cite{chatzikokolakis2015geo}} The principle of geo-indistinguishability (GI)~\cite{andres2013geo}, is a formal notion of privacy that protects the user's exact location with a level of privacy that depends on radius r, which corresponds to a generalized version of differential privacy (DP). GI-DP is a mechanism for achieving geo-indistinguishability when the user releases his location repeatedly throughout the day. It fulfills desired protection level by perturbing the actual location with random noises and achieving an optimal trade-off between privacy and utility (i.e., service quality). We re-implement the geo-indistinguishability of optimal utility with graph spanner~\cite{chatzikokolakis2015geo}, namely GI-DP in this paper, to attain the released version data that satisfied the DP guarantees. We then train a series of Optimal-IMs to evaluate the effectiveness of target inference attacks on the released version data in a line to compare with our proposed mechanism.

\subsection{Training}
\subsubsection{Training of Mo-PAE}
The main goal of the proposed adversarial network is to learn an efficient feature representation based on the utility and privacy budgets, using all users' mobility histories. In most experiments in this work, the trajectory sequences consist of 10 historical locations with timestamps (i.e., $SL=10$), and the impact of the varying sequence lengths is discussed in Section~\ref{section:impact sl}.
After data pre-processing, 80\% of each user's records are segmented as the training set and the remaining 20\% as the testing set. We utilize the mini-batch learning method with the size of 128 to train the model until the expected convergence. We take a gradient step to optimize the \textit{sum loss} $L_{sum}$ (i.e., Equation \ref{equ:sumloss}) in terms of $L_R$, $L_U$, and $L_P$ concurrently. Meanwhile, the \textit{sum loss} $L_{sum}$ is optimized by using the Adam optimizer. All the experiments are performed with the Tesla V100 GPU; a round of training would take 30 seconds on average, and each experiment trains for 1000 rounds.

\subsubsection{Training of the TrajGAN}
To provide a state-of-the-art machine learning-based model for comparison, we re-implement the TrajGAN model described in \cite{rao2020lstm} using the same hyperparameters, setting latent vector dimension to 100, using 100 LSTM units per layer, a batch size of 256, utilizing the Adam optimizer with learning rate 0.001 and momentum 0.5, and training for 200 epochs (where one epoch is a pass through the entire training set). We train TrajGAN independently on the training split of each benchmark mobility dataset, and then use it to generate synthetic trajectories from the test set. Then we train the proposed Mo-PAE on the same training data and use it to generate a feature extraction from the same test data. Finally, we evaluate the performance of the user re-identification unit and mobility prediction unit on the real and synthetic test sets generated by TrajGAN, and compare the changes in accuracy to assess the relative utility and privacy of the TrajGAN and Mo-PAE.

\subsubsection{Training of the DP-GI}
We re-implement the DP-GI model described in~\cite{chatzikokolakis2015geo} using the default settings. That is, we set $epsilon$  $\epsilon=0.5$, $dilation$ $ \delta=1.1$, the \emph{distance matrix} $d_x$ is defined by Euclidean distance. From ~\cite{chatzikokolakis2015geo}, let $X$ be a set of locations with metric $d_x$, and let $G(X, E)$ be a $\delta-spanner$ of $X$, if a mechanism K for X is $\frac\epsilon{\delta}d_G$-private, then K is $\epsilon d_x$-private. The dilation of G is calculated as:
\begin{equation}
 \delta = \underset{x \neq x' \in X}{max} \frac{d_G(x,x')}{d_x(x, x')}
\label{equ:dilation}
\end{equation}
\begin{equation}
 d_G(x, x') \geq d_x (x, x')  \hspace{0.5cm} \forall x, x' \in X
\label{equ:note}
\end{equation}

We re-implement the GI-DP to attain the released version data that satisfied the DP guarantees. We then train a series of Optimal-IMs to evaluate the effectiveness of target inference attacks on the released version data in a line to compare with our proposed mechanism.

\subsection{Metrics}
We set \textit{Euclidean}~\cite{ball1960short} and \textit{Manhattan} distance~\cite{black1998dictionary} as our evaluation metrics for the DRU to evaluate the quality of the reconstructed data $X'$ generated from extracted features $f$. Both distance metrics are widely used in location privacy literature~\cite{andres2013geo, erdemir_privacy-aware_2020}. For instance, the work introducing Geo-Indistinguishability~\cite{andres2013geo} utilizes a privacy level that depends on the \textit{Euclidean} distance. \textit{Euclidean} distance gives the shortest or minimum distance between two points. In contrast, \textit{Manhattan} distance applies only if the points are arranged in a grid, and both definitions are feasible for the problem we are working on. Note that these two distances have limited capability in showing the quality of the reconstructed data $X'$, however, they intuitively capture the differences between the original data $X$ and the reconstructed data $X'$. 

For both MPU and URU, we leverage the top-n accuracy as our evaluation metric. The accuracy of the MPU is one of the most important factors in evaluating the utility of the extracted feature representation \textit{f}, where predictability of the \textit{f} increases as much as it can during the adversarial training. On the other hand, the competing training objective is to decrease the accuracy of the user re-identification unit to enhance the privacy of \textit{f}. The top-n metric computes the number of times the correct label appears among the predicted top $n$ labels. The top-n metric takes n predictions with higher probability into consideration, and it classifies the prediction as correct if one of them is an accurate label. The top-1 to top-5 accuracies are leveraged in our paper to discuss the performance of the proposed model.

\begin{table*}[t!]
   \centering
   \begin{tabular}{cccccccccccc}
    \hline
    \multirow{2}{*}{\shortstack{Datasets}} & \multirow{2}{*}{\shortstack{Models}} & & \multicolumn{2}{|c|}{Privacy I} & \multicolumn{3}{c|}{Utility (\% for loss)} & \multicolumn{3}{c|}{Privacy II (\% for gain)} & Utility-PII\\
    \cline{4-11}
        &  & & \multicolumn{1}{|c}{Euc} & \multicolumn{1}{c|}{Man} & top-1  & top-3  & \multicolumn{1}{c|}{top-5} & top-1  & top-3 & \multicolumn{1}{c|}{top-5} & trade-offs (\%) \\
        \hline
        
        \multirow{5}{*}{\shortstack{MDC}} & Optimal-IMs & & 0.0000 & 0.0000 & 0.9347  & 0.9837  & 0.9922 & 0.9247 & 0.9819 & 0.9911 & -\\
        & TrajGAN & & 0.0434 & 3.6923  & -46.32\% & -24.16\% & -15.98\% & +20.32\% & +8.13\% & +4.02\% & -26.00\%\\
        & GI-DP & & {\bf 0.2341} & {\bf 56.8764} &  -97.34\% & -93.25\% & -89.44\% & {\bf +97.47\%} & {\bf +93.71\%} & {\bf +90.12\%} & 0.13\% \\
        & \multirow{2}{*}{\shortstack{Our Model}} & I & 0.0025 & 0.4501 & -54.56\% & -34.74\% & -25.10\% & +69.80\% & +50.44\% & +39.95\% & 15.24\%\\
        & & II & 0.0697 & 13.6168 & \textbf{-13.43\%} & \textbf{-6.26\%} & \textbf{-3.95\%} & +65.51\% & +45.11\% & +34.86\% & {\bf 52.08\%}\\
        \hline
        
        \multirow{5}{*}{\shortstack{Priva'Mov}} & Optimal-IMs & & 0.0003 & 0.0058 & 0.9482  & 0.9878 & 0.9954 & 0.5643 & 0.8215 & 0.8765 & - \\
        & TrajGAN & & 0.0815 & 9.6843 & -6.60\% & -1.89\% & -0.93\% & +14.17\% & +14.35\% & +8.88\% & 7.57\%\\
        & GI-DP & & 0.1899 & {\bf 38.6712} &  -91.20\% & -83.53\% & -72.37\% & {\bf +85.49\%} & {\bf +63.80\%} & {\bf +53.31\%} & -5.71\%\\
        & \multirow{2}{*}{\shortstack{Our Model}} & I & 0.0009 & 0.0437 & \textbf{-3.36\%} & \textbf{-1.59\%} & \textbf{-0.81\%} & +27.02\% & +14.19\% & +9.19\% & 23.66\%\\
        & & II & \textbf{0.2347} & 10.2239 & -10.81\% & -6.83\% & -4.91\% & +35.29\% & +14.97\% & +10.05\% & {\bf 24.48\%}\\
        \hline
        
        \multirow{5}{*}{\shortstack{Geolife}} & Optimal-IMs & & 0.0008 & 0.0670 & 0.4705 & 0.6842 & 0.7636 & 0.6572 & 0.8690 & 0.9294 & -\\
        & TrajGAN &  & 0.4010 & 50.3620 & -62.31\% & -50.45\% & -43.72\% & +66.73\% & +47.89\% & +37.22\% & 4.42\%\\
        & GI-DP & & {\bf 1.2332} & {\bf 312.9972} &  -97.74\% & -96.56\% & -95.36\% & {\bf +91.57\%} & {\bf +84.13\%} & {\bf +78.65\%} & -6.17\%\\
        & \multirow{2}{*}{\shortstack{Our Model}} & I & 0.0006 & 0.0310 & -31.45\% & -25.02\% & -21.90\% & +54.88\% & +39.59\% & +30.81\% & 23.43\%\\
        & & II & 0.4351 & 89.2209 & \textbf{-21.13\%} & \textbf{-18.78\%} & \textbf{-17.11\%} & +55.49\% & +40.40\% & +32.34\% & {\bf 34.36\%}\\
        
        \hline
        \multirow{5}{*}{\shortstack{FourSquare}} & Optimal-IMs & & 0.0052 & 0.6691 & 0.6468 & 0.8210 & 0.8823 & 0.8780 & 0.9735 & 0.9892 & - \\
        & TrajGAN & & \textbf{1.4341} & \textbf{117.9181} & -26.30\% & -22.30\% & -18.75\% & +51.86\% & +32.49\% & +23.49\% & 25.56\%\\
        & GI-DP & & 0.5826 & 86.096 &  -69.35\% & -59.23\% & -53.36\% & {\bf +77.29\%} & {\bf +66.58\%} & {\bf +59.82\%} & 7.94\%\\
        & \multirow{2}{*}{\shortstack{Our Model}} & I & 0.0060 & 0.7845 & -51.05\% & -41.45\% & -35.20\% & +53.47\% & +35.26\% & +25.86\% & 2.42\%\\
        & & II & 0.7985 & 99.9212 & \textbf{-2.54\%} & \textbf{-3.14\%} & \textbf{-2.84\%} & +51.08\% & +34.39\% & +26.16\% & {\bf 48.54\%}\\
        \hline
        
    \end{tabular}
    \caption{Performance comparison between Mo-PAE with other baseline models. The \textit{Model I} is our proposed architecture without weights, and the \textit{Model II} is the one with multipliers ($\lambda_1$ = 0.1, $\lambda_2$ = 0.8, and $\lambda_3$ = 0.1). The results shown in this table are all with trace sequence length 10 (i.e., \textit{SL} = 10). 
    %The similarity indexes (Euclidean and Manhattan distance) are leveraged to intuitively represent the difference between the original data and reconstructed data. 
    %The \textit{Utility} and \textit{Privacy} columns of Standalone Model represent the upper bound of predictability and user re-identification capability of the separated units without consideration of utility-privacy tradeoffs, respectively. 
    The \textit{Privacy I} intuitively shows the difference between the raw data and reconstructed data; the \textit{Utility} (\%) represents the utility loss; and the \textit{Privacy II}(\%) represents the privacy gain calculated via the decline of the user re-identification accuracy. 
    %Note that the user re-identification accuracy is leveraged as the privacy metric here, and the privacy gain is calculated by the user re-identification  \textit{error}.
    }
    \label{tab:compare}
\end{table*}

\section{Architecture Evaluation}
\label{FraE}

In this section, we present the comparison results between the proposed Mo-PAE and three baseline models under the same training setting. 

\subsection{Performance Comparison}
We first compare our proposed models with the Optimal-IMs, TrajGAN, and DP-GI on four representative mobility datasets, as shown in Table~\ref{tab:compare}. The overall performance is evaluated in terms of the \textit{utility level} provided by the MPU and the \textit{privacy protection} provided by DRU and URU. The \textit{Model I} is our proposed architecture but without applying the Lagrange multipliers(i.e., where each unit is weighted equally). The \textit{Model II} is the one with Lagrange multipliers (i.e., $\lambda_1$, $\lambda_2$, $\lambda_3$ ) and different weights are given to units (i.e., $\lambda_1=0.1,\ \lambda_2=0.8,\ \lambda_3=0.1$ for the results in Table~\ref{tab:compare}). In this table, the sequence length of the input traces is 10, that is $SL=10$. We will discuss why we choose SL=10 and the impact of the SL in Section \ref{sec:discussions}.

As we mention in Section~\ref{sec:baseline}, Optimal-IMs are trained without considering the privacy-utility trade-offs; hence, they can be leveraged to explain the optimal inference accuracy achieved. That is, before any privacy-preserving mechanism applies, the accuracy of the target or private tasks with raw data. For instance, the accuracy of the \emph{Privacy II} (0.9247 (MDC), 0.5643 (Priva'Mov), 0.6572 (GeoLife), 0.8780 (FourSquare)) demonstrates that an adversary can accurately infer user identity from raw data before any privacy protection.

Different from Optimal-IMs, the other models consider the privacy-utility trade-offs, and we measure the privacy protection and data utility by the effectiveness of the inference units. Therefore, the results in Table~\ref{tab:compare} are shown in \emph{utility loss} and \emph{privacy gain}, both of which are in a percentage format (\%), compared with the accuracy of Optimal-IMs. To compare the trade-off between them more intuitively, we list the "Utility-PII trade-off" column, where "trade-offs = Utility (\% for loss) + Privacy II (\% for gain)". The distance indexes (i.e., \emph{"Euc"} and \emph{"Man"}) are leveraged to intuitively represent the difference between the original data $X$ and reconstructed data $X'$, where a larger value indicates numerical differences between $X$ and $X'$. For the distance index, we are interested in the distance between each trace, hence, we consider the quantity of trace for datasets $X_D$ and get these distance indexes by averaging the corresponding record numbers $N_D$, that is (take \emph{"Euc"} for example):
\begin{equation}
    Euc (X_D, X_D') =  \frac{ \sqrt{\sum_{i=1}^{N_D} (x_i -x_i')^2}}{N_D}, \hspace{0.5cm} N_D = N_D'
\label{eq:euc}
\end{equation}

\begin{figure*}
     \centering
     \begin{subfigure}[t]{0.8\textwidth}
     \centering
         \includegraphics[width=\textwidth]{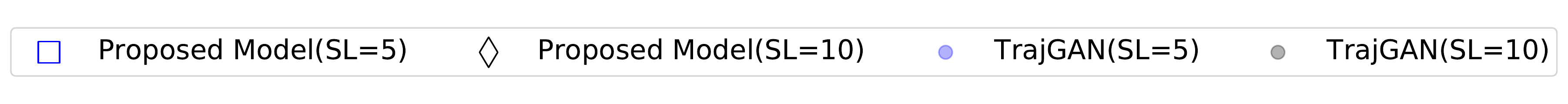}
     \end{subfigure}
     
     \begin{subfigure}[t]{0.24\textwidth}
         \centering
         \includegraphics[width=\textwidth]{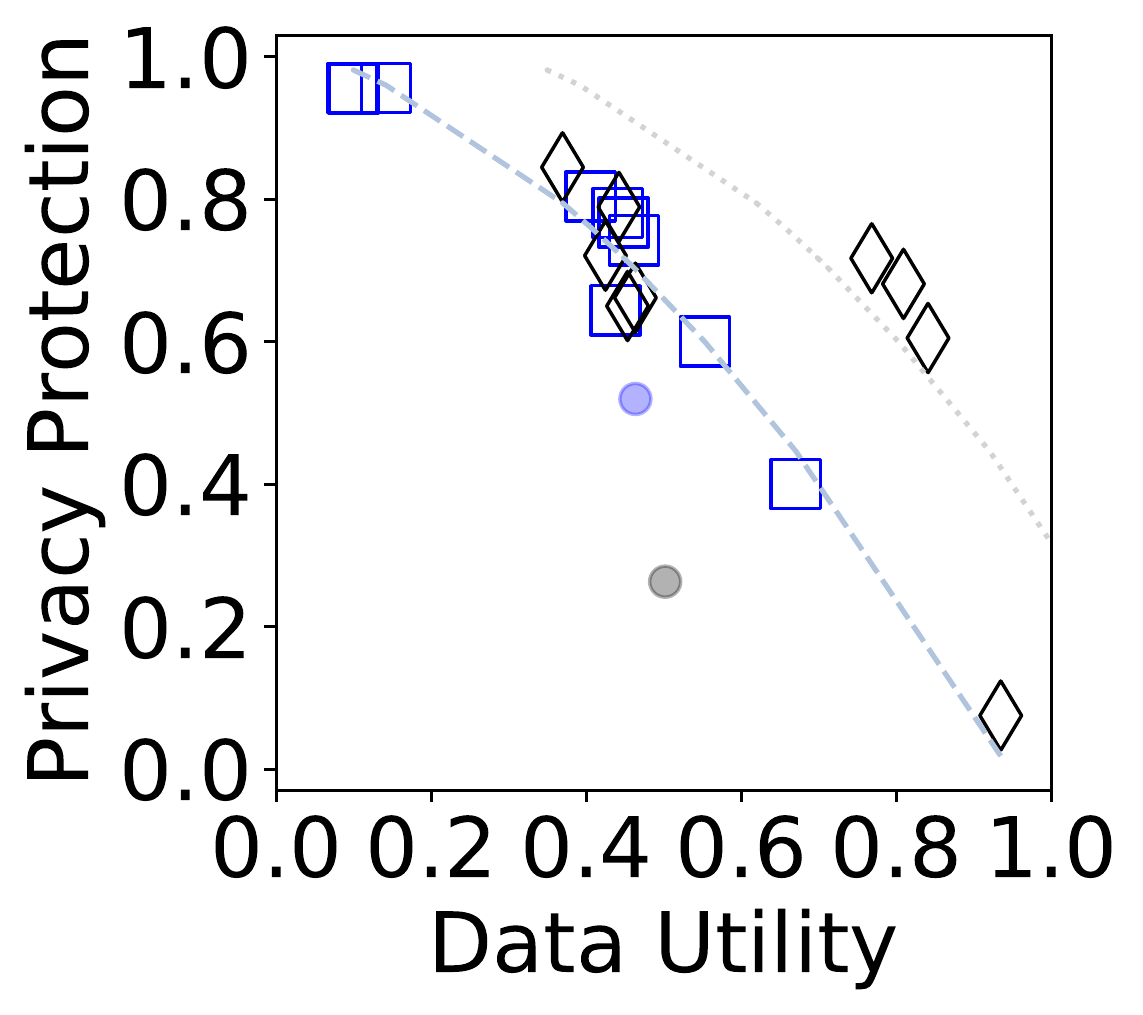}
         \caption{MDC}
         \label{fig:mdc}
     \end{subfigure}
     \begin{subfigure}[t]{0.24\textwidth}
         \centering
         \includegraphics[width=\textwidth]{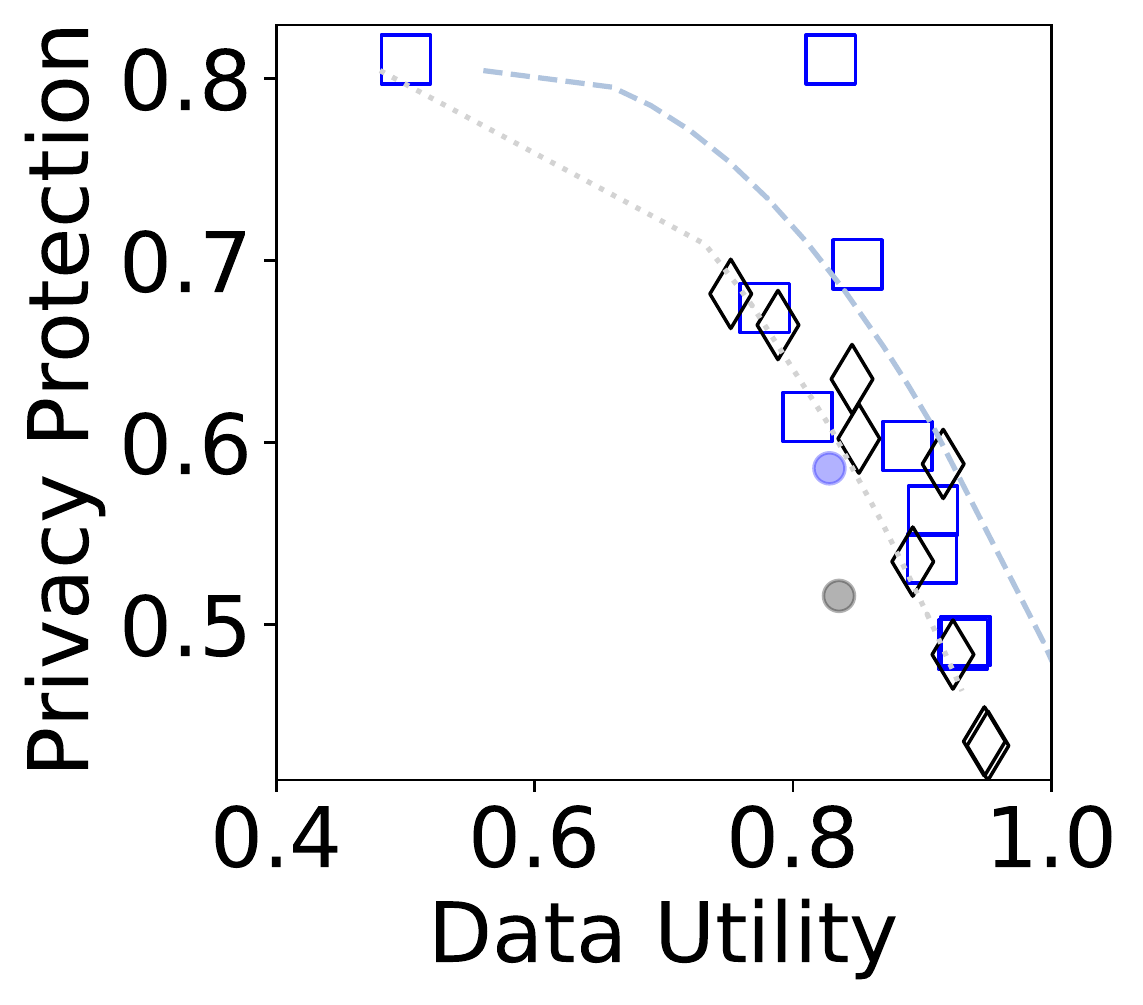}
         \caption{Priva\'Mov}
         \label{fig:privamov}
    \end{subfigure}
    \begin{subfigure}[t]{0.23\textwidth}
         \centering
         \includegraphics[width=\textwidth]{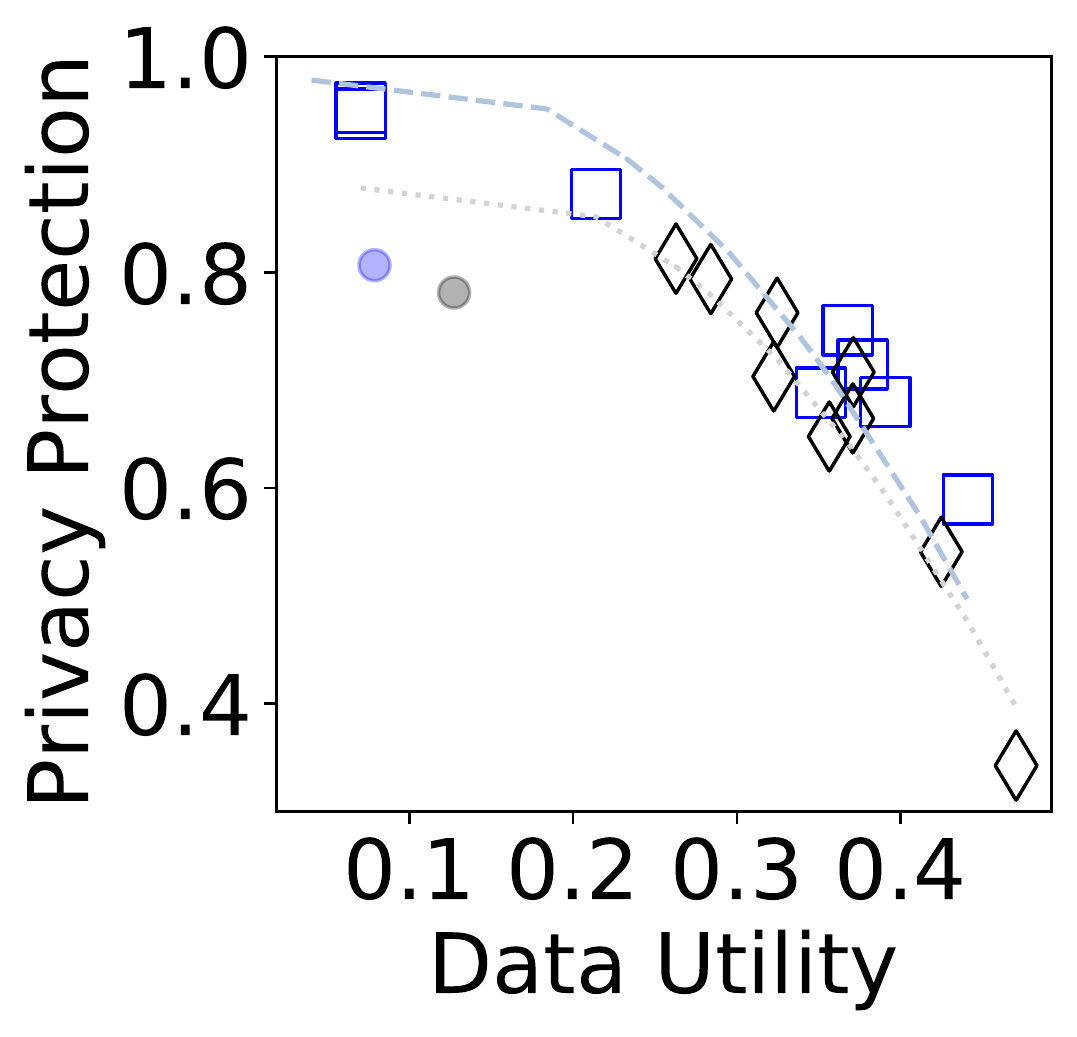}
         \caption{Geolife}
         \label{fig:geolife}
     \end{subfigure}
     \begin{subfigure}[t]{0.23\textwidth}
         \centering
         \includegraphics[width=\textwidth]{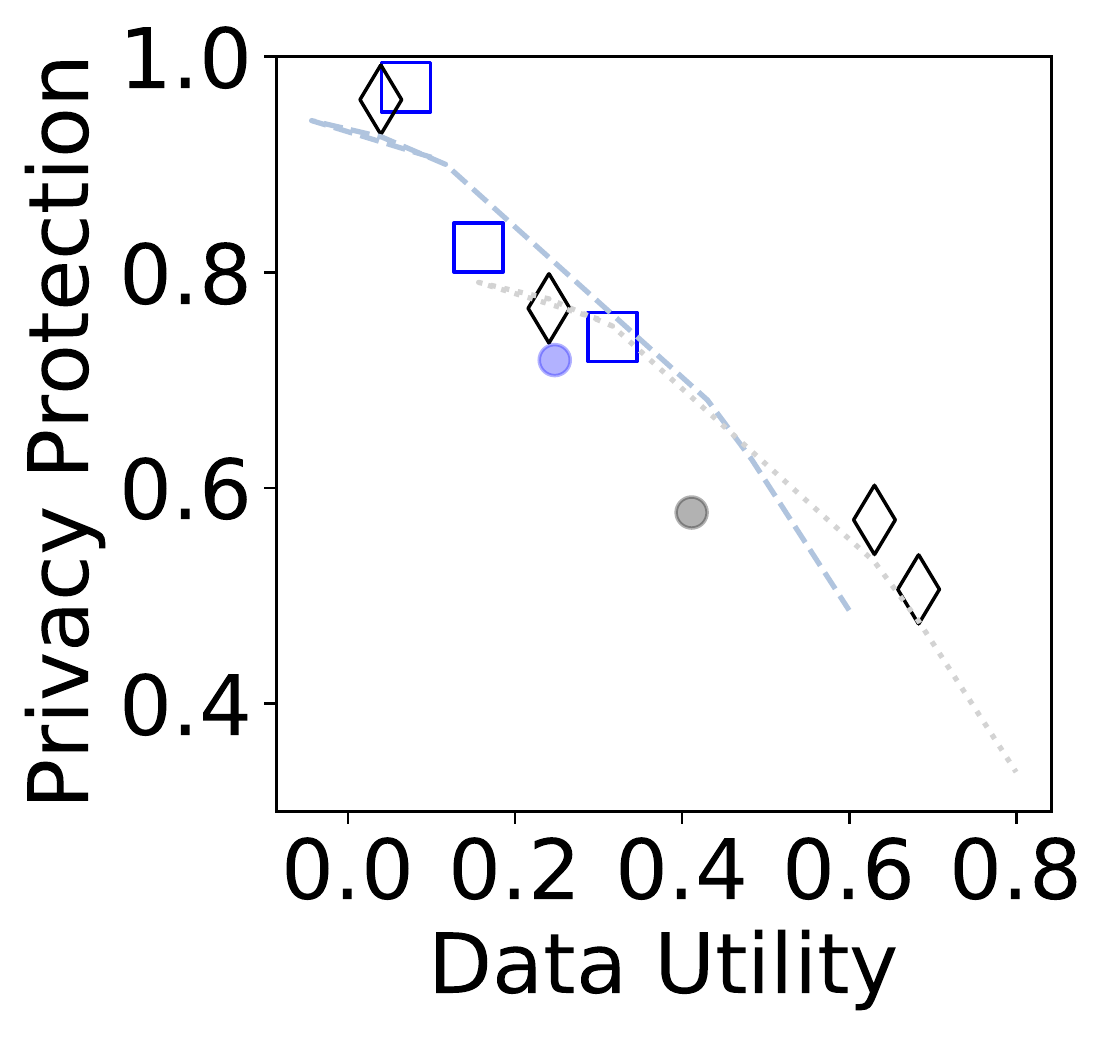}
         \caption{Foursquare}
         \label{fig:foursquare}
    \end{subfigure}
    \caption{Pareto Frontier trade-off analysis on four datasets. The hollow squares and diamonds present the results of the proposed models Mo-PAE.
    solid points present the results of the TrajGAN.
    Blue color means \textit{SL} = 5. Black color means \textit{SL} = 10.}
    %\Description{}
    \label{fig:compare}
\end{figure*}

\begin{table*}[t]
   \centering
   \begin{tabular}{ccccccccccccc}
    \hline
    \multicolumn{5}{c}{Settings} & \multicolumn{2}{|c|}{MDC} & \multicolumn{2}{c|}{Priva'Mov} & \multicolumn{2}{c|}{Geolife} & \multicolumn{2}{c}{FourSquare} \\
    \cline{1-13}
         & & $\lambda_1$  &  $\lambda_2$ &  $\lambda_3$  &  \multicolumn{1}{|c}{Euc}  &  \multicolumn{1}{c|}{Utility}  &  Euc  &  \multicolumn{1}{c|}{Utility}  &  Euc  &  \multicolumn{1}{c|}{Utility}  &  Euc  &  Utility \\
        \hline
        Model I & - & -   & -   & - & \multicolumn{1}{|c}{+0.0017} & -30.27\% & +0.0009 & -2.72\%  & +0.0057 & -17.9\%  & +0.0069 & -33.75\% \\
       \cline{1-5}
        \multirow{5}{*}{\shortstack{Model II}} 
        & i   & 0.1 & 0.8 & 0.1 & \multicolumn{1}{|c}{+0.0697} & -12.55\% & +0.0453 & -2.71\%  & +0.4343 & -9.94\%  & +0.7933 & -1.64\%  \\
        & ii  & 0.2 & 0.6 & 0.2 & \multicolumn{1}{|c}{+0.0791} & -33.29\% & +0.0738 & -10.72\% & +0.4889 & -18.21\% & +1.2722 & -50.50\% \\
        & iii & 0.3 & 0.4 & 0.3 & \multicolumn{1}{|c}{+0.0889} & -58.10\% & +0.0782 & -16.56\% & +0.5220 & -29.95\% & +1.9586 & -60.71\% \\
        & iv  & 0.1 & 0.6 & 0.3 & \multicolumn{1}{|c}{+0.0822} & -49.27\% & +0.0776 & -10.28\% & +0.4717 & -18.64\% & +1.4139 & -57.40\% \\
        \hline
    \end{tabular}
    \caption{Impact of Mo-PAE on the data reconstruction accuracy (\emph{PI}) and relative utility loss (\textit{U}) on four mobility datasets. 
    We list \emph{Model I} and four different settings of \emph{Model II}'s weight combinations to discuss the potential range of the trade-offs.}
    \label{tab:euc}
\end{table*}

Table~\ref{tab:compare} demonstrates that our proposed models, especially \emph{Model II}, outperform the TrajGAN and GI-DP across various datasets. For instance, with the MDC dataset, our \textit{Model II} achieves the best trade-offs when compared with other models, as the utility loss is only 13.43\% but with 65.51\% privacy gain, while 46.32\% utility loss and 20.32\% privacy gain with the TrajGAN, and 97.34\% utility loss and 97.47\% privacy gain with the GI-DP. The extreme performance on the GI-DP illustrates that while the DP paradigm is a robust privacy-preserving technique in protecting \emph{user's location}, it is not appropriate in protecting the \emph{user's identities}.

More intuitively, in the column of "trade-off", \textit{Model II} achieves all the best trade-offs among four datasets (52.08\% (MDC), 24.48\% (Priva'Mov), 34.36\% (GeoLife) and 48.54 (FourSquare)).
\emph{Model I} has worse performance than \emph{Model II}, in general, but is still superior to TrajGAN and DP-GI, where the latter two might even get \emph{negative} trade-offs (i.e., TrajGAN got -26.00\% with MDC and GI-DP got -5.71\% with Priva'Mov). Moreover, for the Priva'Mov dataset, although the utility loss of the TrajGAN is 4.21\% smaller than our \textit{Model II}, both two privacy metrics of the TrajGAN are worse than the \textit{Model II}. Again, our model has better overall trade-offs, as 23.66\% for \textit{Model I} and 24.48\% for \textit{Model II}. The performance on Geolife and FourSquare are similar but inverse, where the utility of our model is better than TrajGAN and with slightly weaker privacy preservation. 

In summary, GI-DP always has the highest privacy gain among the four datasets, however, the utility loss is also very high, resulting in inadequate and unexpected privacy-utility trade-offs. This trend also shows that the DP mechanism is not an appropriate metric for the location privacy of \emph{user's identity}, which is also in line with the conclusions from other related work~\cite{shokri2011quantifying, erdemir_privacy-aware_2020}. The comparisons between \textit{Model I} and \textit{Model II} also illustrate the importance of the Lagrange multipliers, which provides flexibility to our proposed architecture, enabling its application in different scenarios and enhancing the privacy-utility trade-offs in this case.

\begin{figure*}[t]
     \centering
     \includegraphics[width=0.99\textwidth]{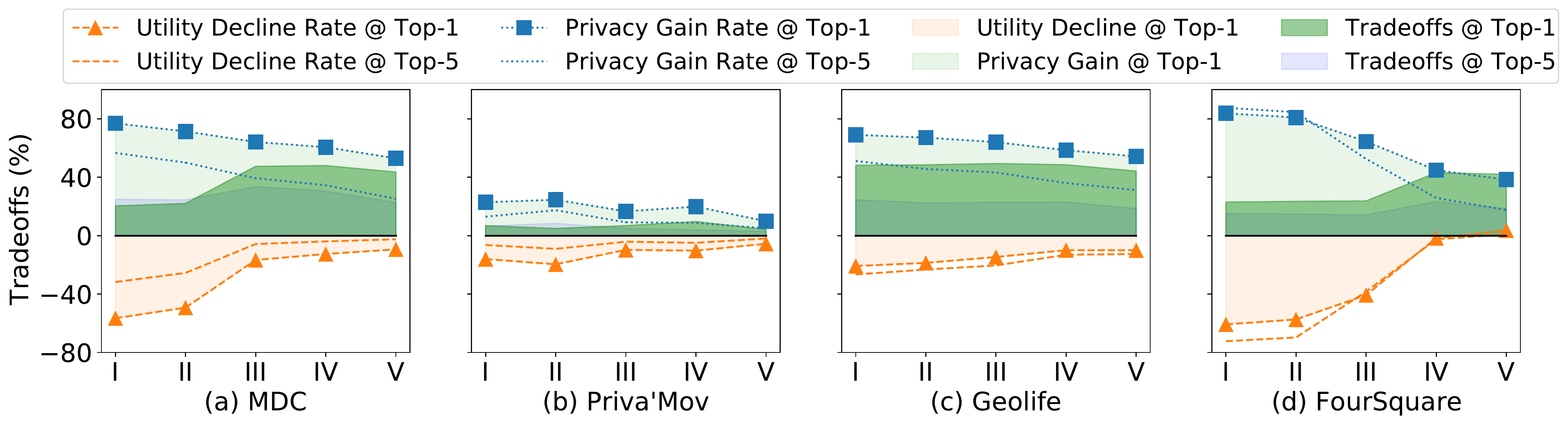}
     \caption{Impact of Mo-PAE on the user re-identification accuracy (PII) and relative utility loss (\textit{U}) on four datasets. 
     The orange area represents the utility loss while the light-green area represents privacy gain. The dark-green area represents the trade-offs between utility achievement and privacy budgets. The x-axis shows five different model settings, and the y-axis shows the trade-offs.}
    %\Description{}
    \label{fig:utility-privacy}
\end{figure*}

\subsection{Trade-off Comparison}

In this section, we present the privacy-utility trade-off analysis between the proposed Mo-PAE and TrajGAN in terms of mobility prediction accuracy (i.e., \textit{U}) and user de-identification efficiency (i.e., \textit{PII}). Figure~\ref{fig:compare} presents the trade-off comparisons of the four datasets, where the \textit{hollow squares} and \textit{hollow diamonds} show the trade-offs provided by the proposed Mo-PAE in $SL=5$ and $SL=10$, respectively. The \textit{solid points} present the results of the TrajGAN under the same experimental setting. As can be seen from these results, in all four cases, the synthetic dataset generated by the TrajGAN is not {\em Pareto-optimal}. That is, the proposed Mo-PAE is able to achieve a better privacy level for a dataset with the same utility value. Compared with the TrajGAN, Mo-PAE improves utility and privacy simultaneously on four datasets. Especially for the performance of MDC, the privacy improves 45.21\% more than the TrajGAN, while the utility also increases by 32.89\%. These results illustrate that our proposed model achieves promising performance in training a privacy-sensitive encoder $Enc_L$ for different datasets.

After evaluating the superior performance of our proposed model, we discuss the privacy guarantee that Mo-PAE provided in terms of data reconstruction(\textit{PI}, \textit{"Euc"} in Table~\ref{tab:euc}) and user re-identification(\textit{PII}, \textit{privacy gain} in Figure~\ref{fig:utility-privacy}). As we mentioned in the Section~\ref{RelW}, the privacy guarantee of Mo-PAE differs from that of DP paradigms and is given in the declined effectiveness of inference attacks.

\begin{figure*}[t]
     \centering
     \begin{subfigure}[b]{0.98\textwidth}
     \centering
         \includegraphics[width=\textwidth]{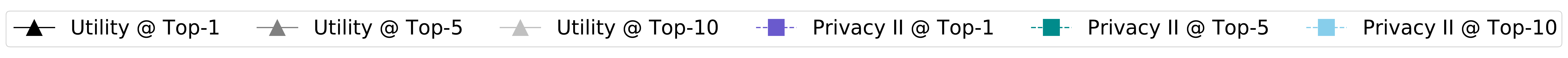}
     \end{subfigure}
     
     \hfill
     \begin{subfigure}[b]{0.24\textwidth}
         \centering
         \includegraphics[width=\textwidth]{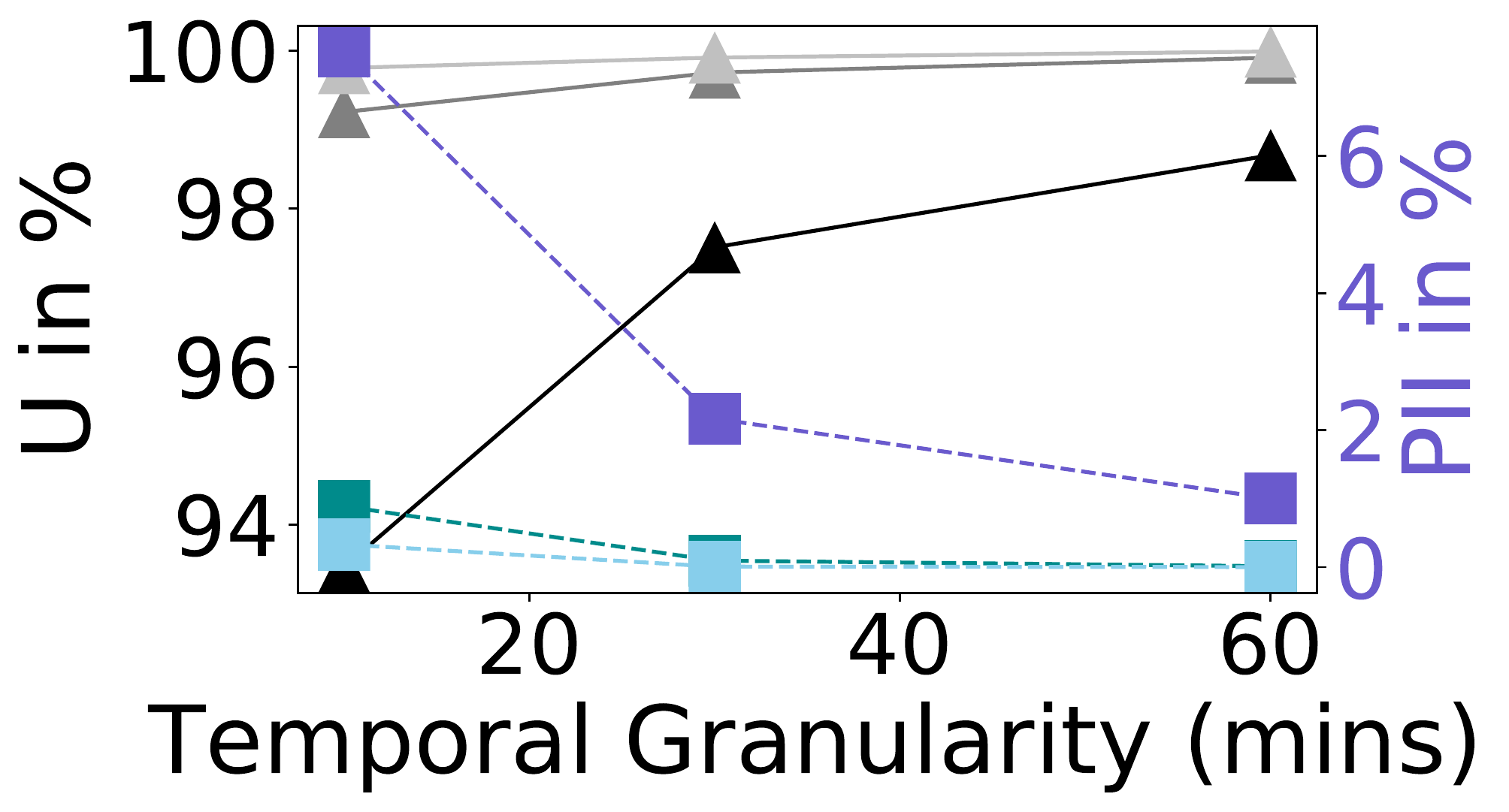}
         \caption{MDC}
         \label{fig:tem_mdc}
     \end{subfigure}
     \hfill
     \begin{subfigure}[b]{0.24\textwidth}
         \centering
         \includegraphics[width=\textwidth]{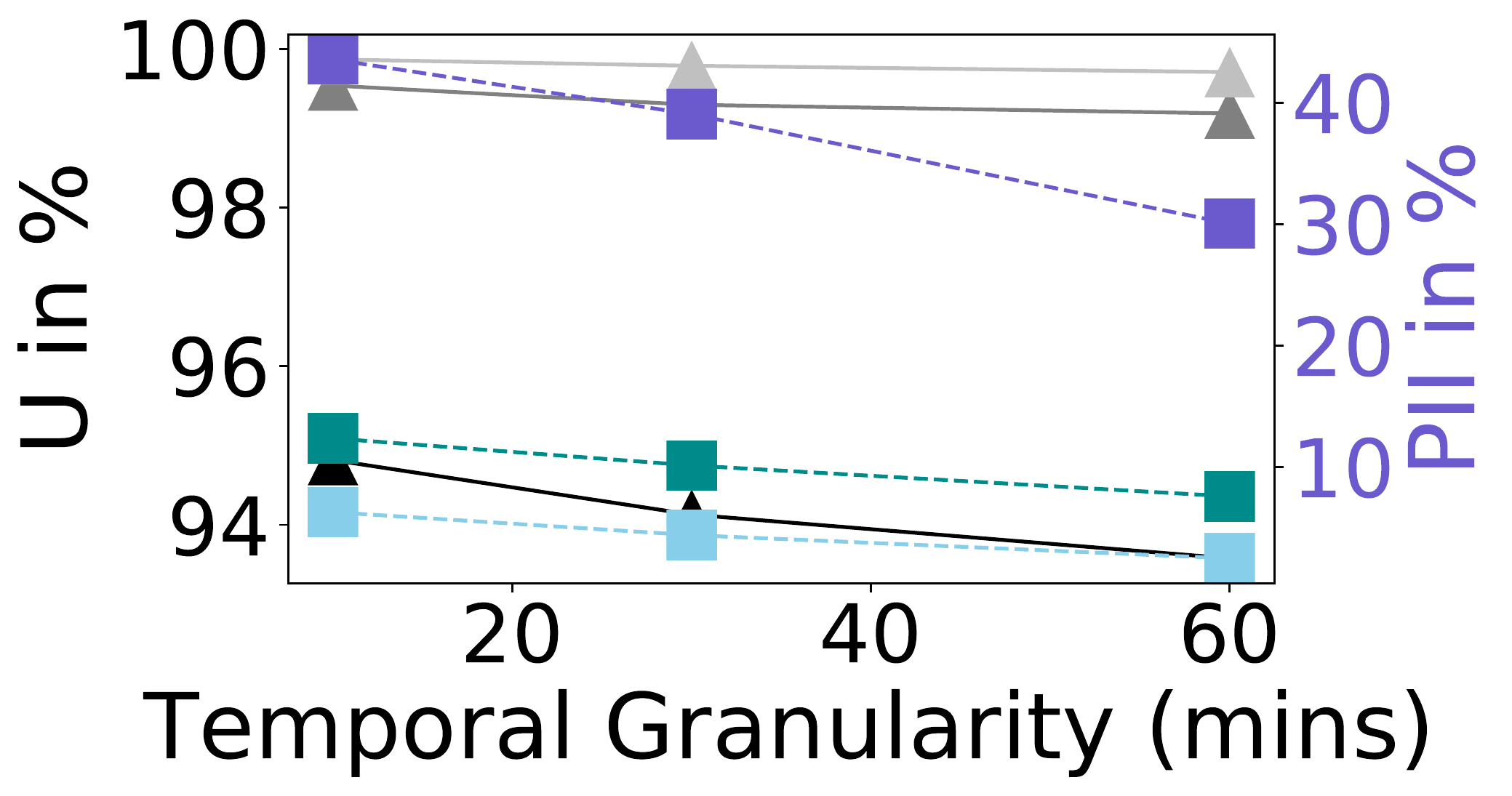}
         \caption{Priva\'Mov}
         \label{fig:tem_privamov}
    \end{subfigure}
    \hfill
    \begin{subfigure}[b]{0.24\textwidth}
         \centering
         \includegraphics[width=\textwidth]{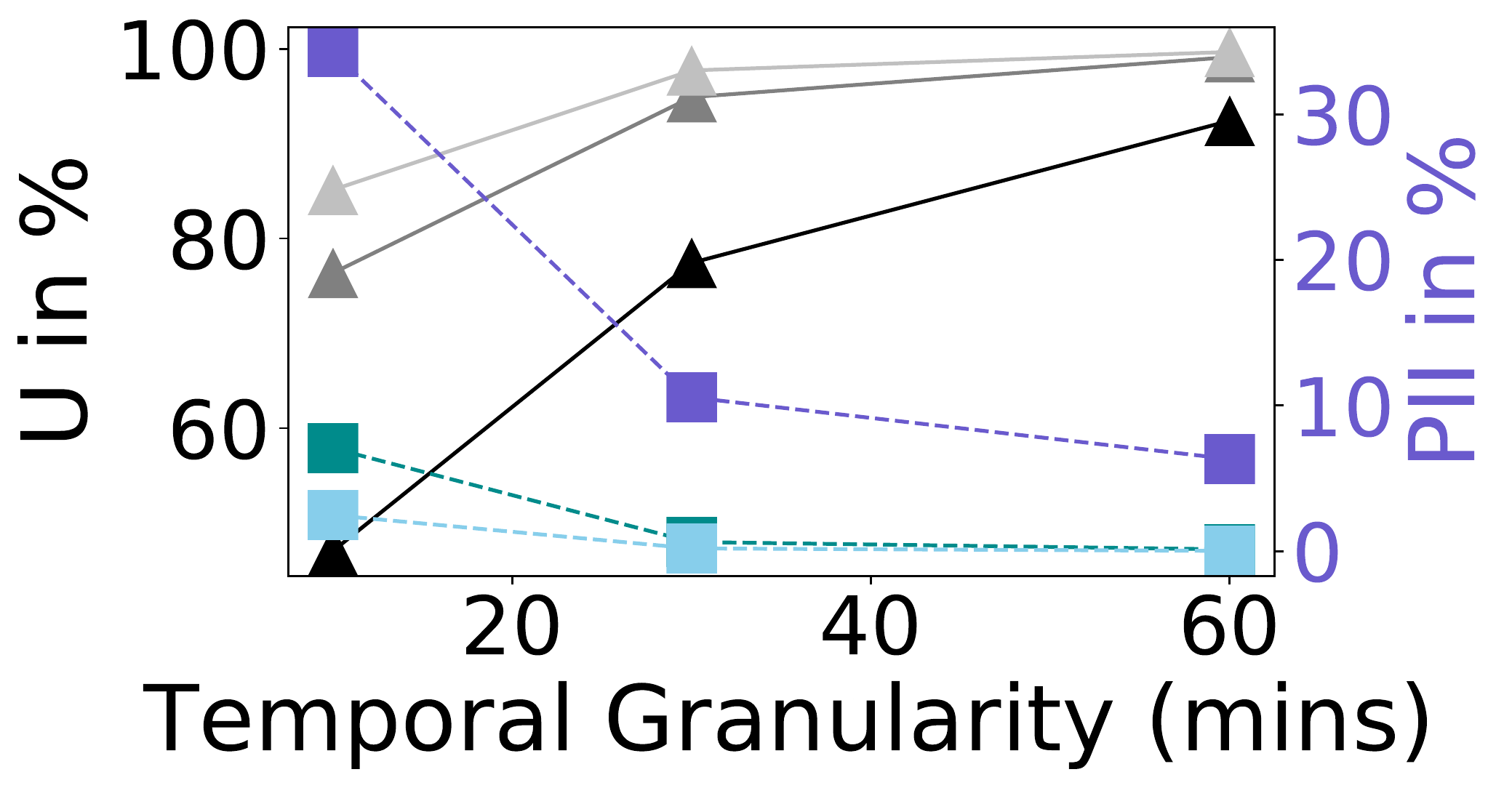}
         \caption{Geolife}
         \label{fig:tem_geolife}
     \end{subfigure}
     \hfill
     \begin{subfigure}[b]{0.24\textwidth}
         \centering
         \includegraphics[width=\textwidth]{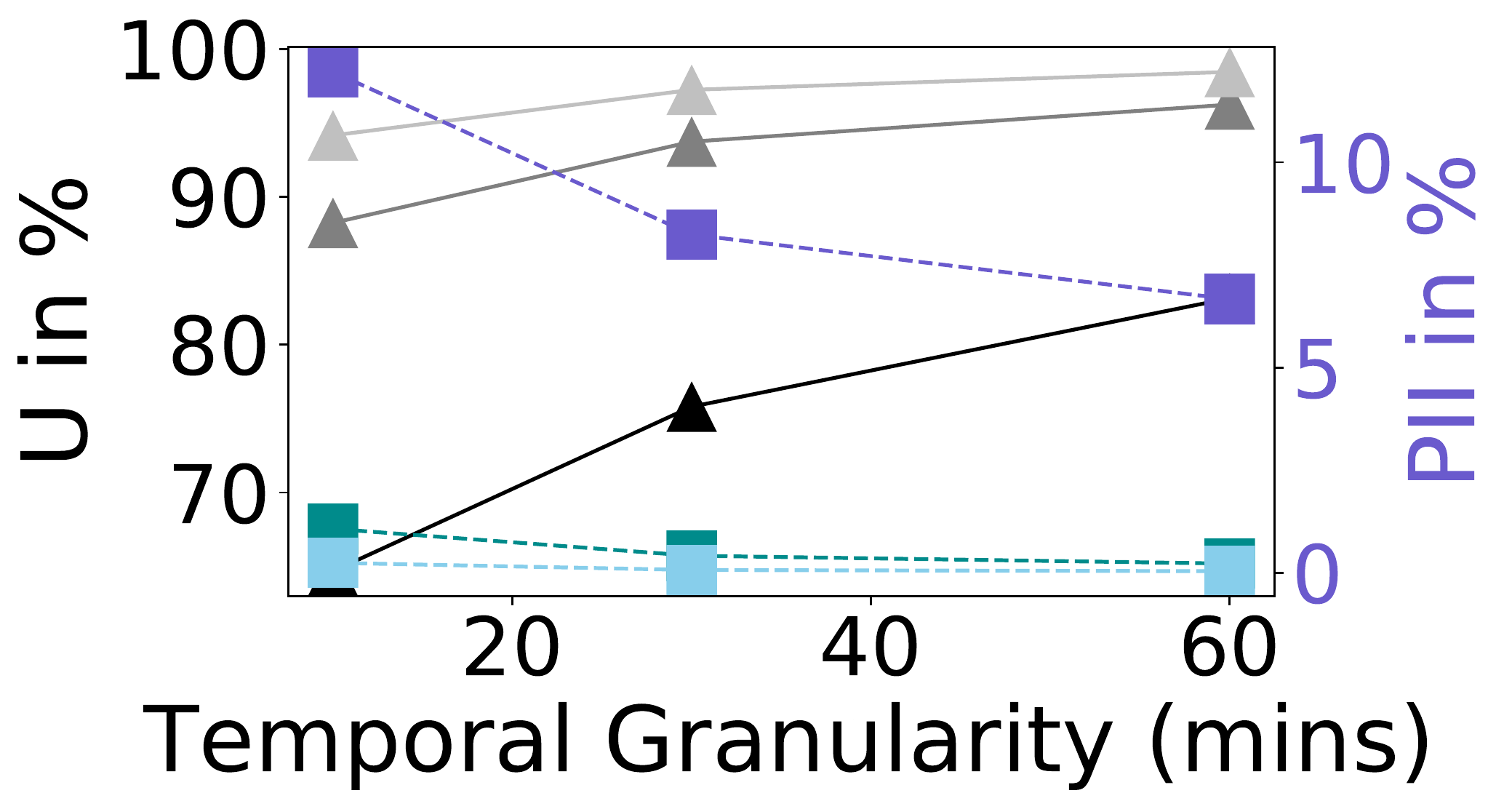}
         \caption{Foursquare}
         \label{fig:tem_foursquare}
    \end{subfigure}
    \hfill
    
    \caption{The effect of temporal granularity on the model performance of four mobility datasets.}
    %\Description{}
    \label{fig:temporal}
\end{figure*}

\begin{figure*}
     \centering
     \hfill
     \begin{subfigure}[b]{0.24\textwidth}
         \centering
         \includegraphics[width=\textwidth]{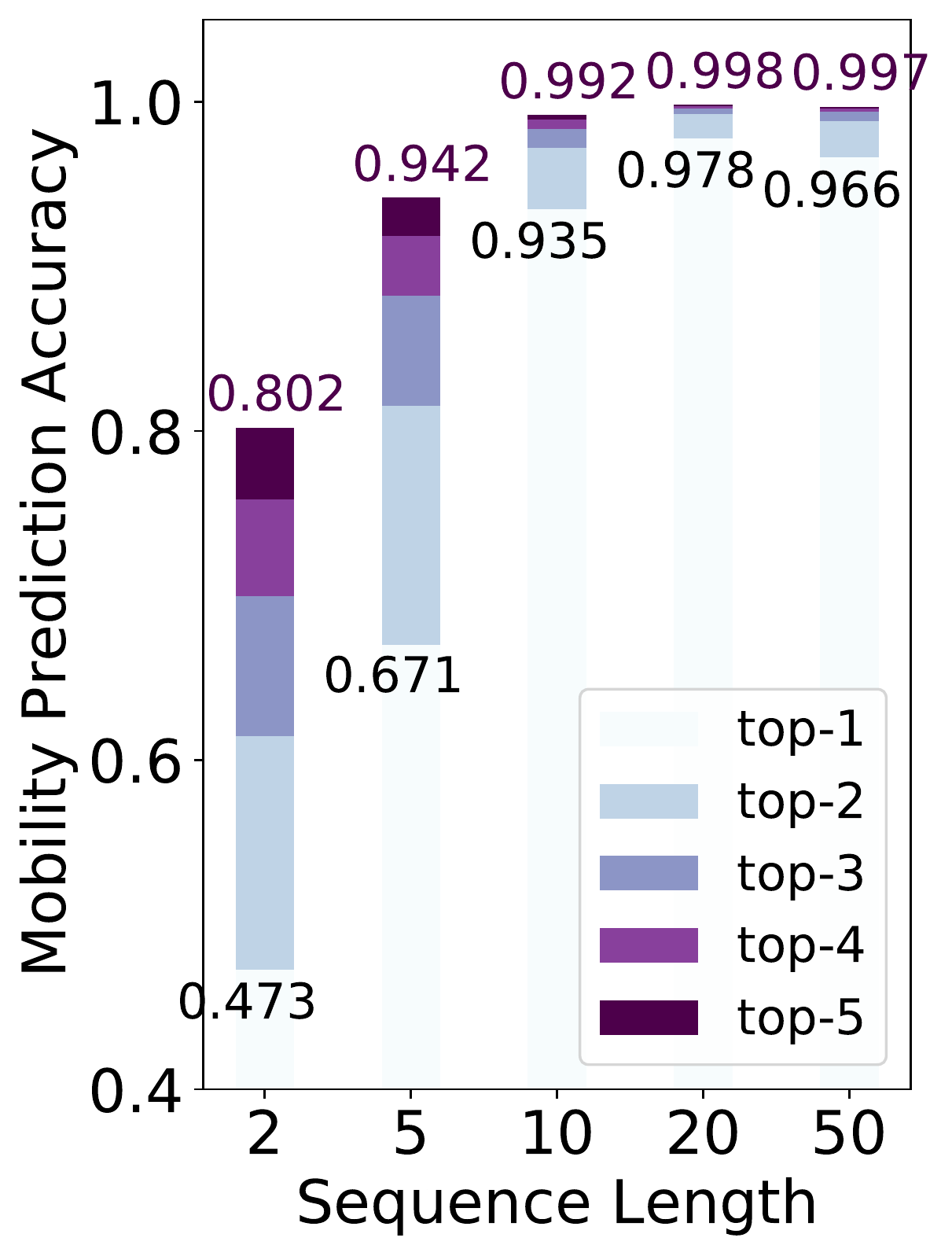}
         \caption{$U_D$ Performance on MDC}
         \label{fig:context_mdc_prediction}
     \end{subfigure}
     \hfill
     \begin{subfigure}[b]{0.24\textwidth}
         \centering
         \includegraphics[width=\textwidth]{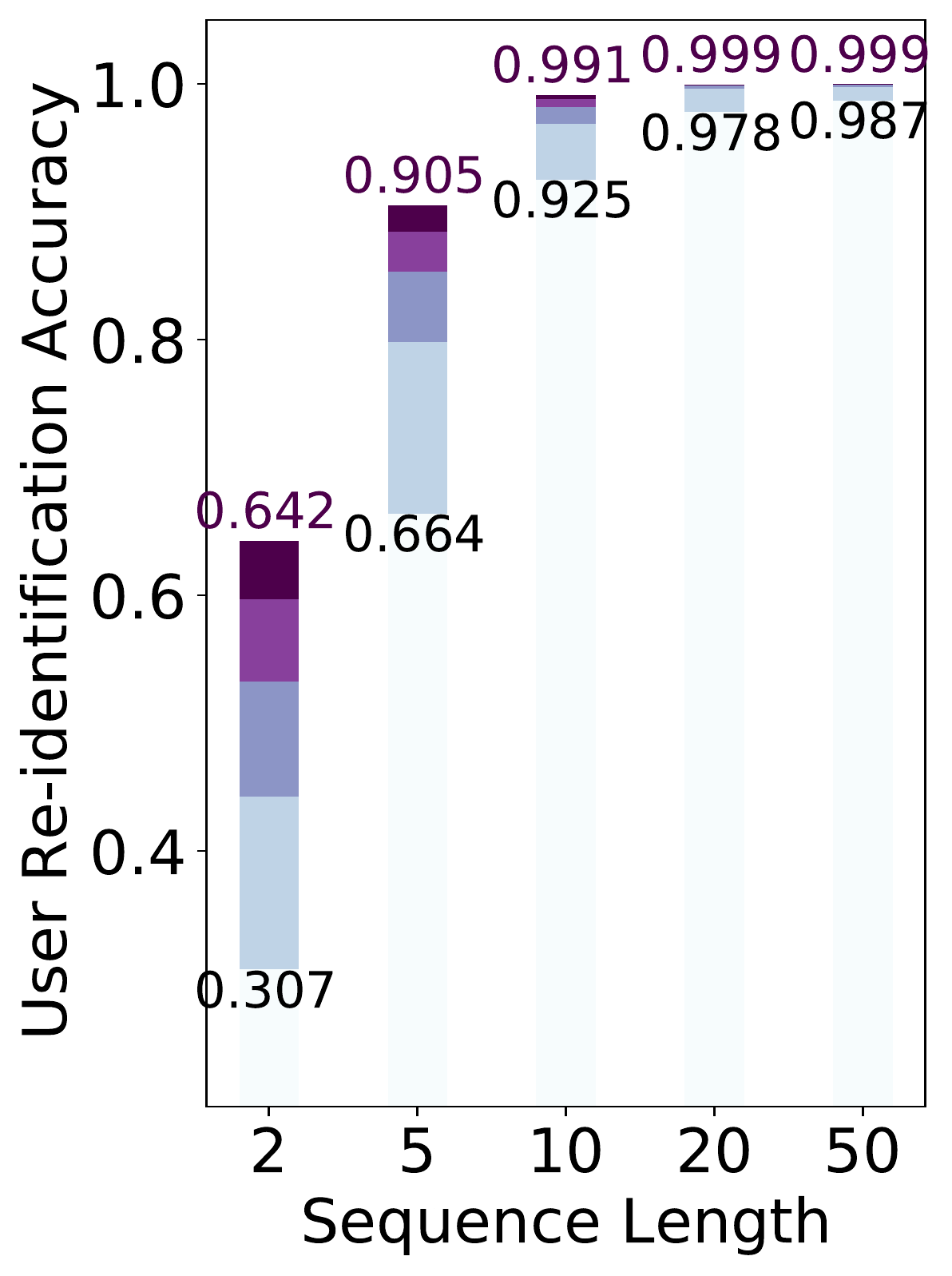}
         \caption{$P_D^2$ Performance on MDC}
         \label{fig:context_mdc_recognition}
    \end{subfigure}
    \hfill
    \begin{subfigure}[b]{0.24\textwidth}
         \centering
         \includegraphics[width=\textwidth]{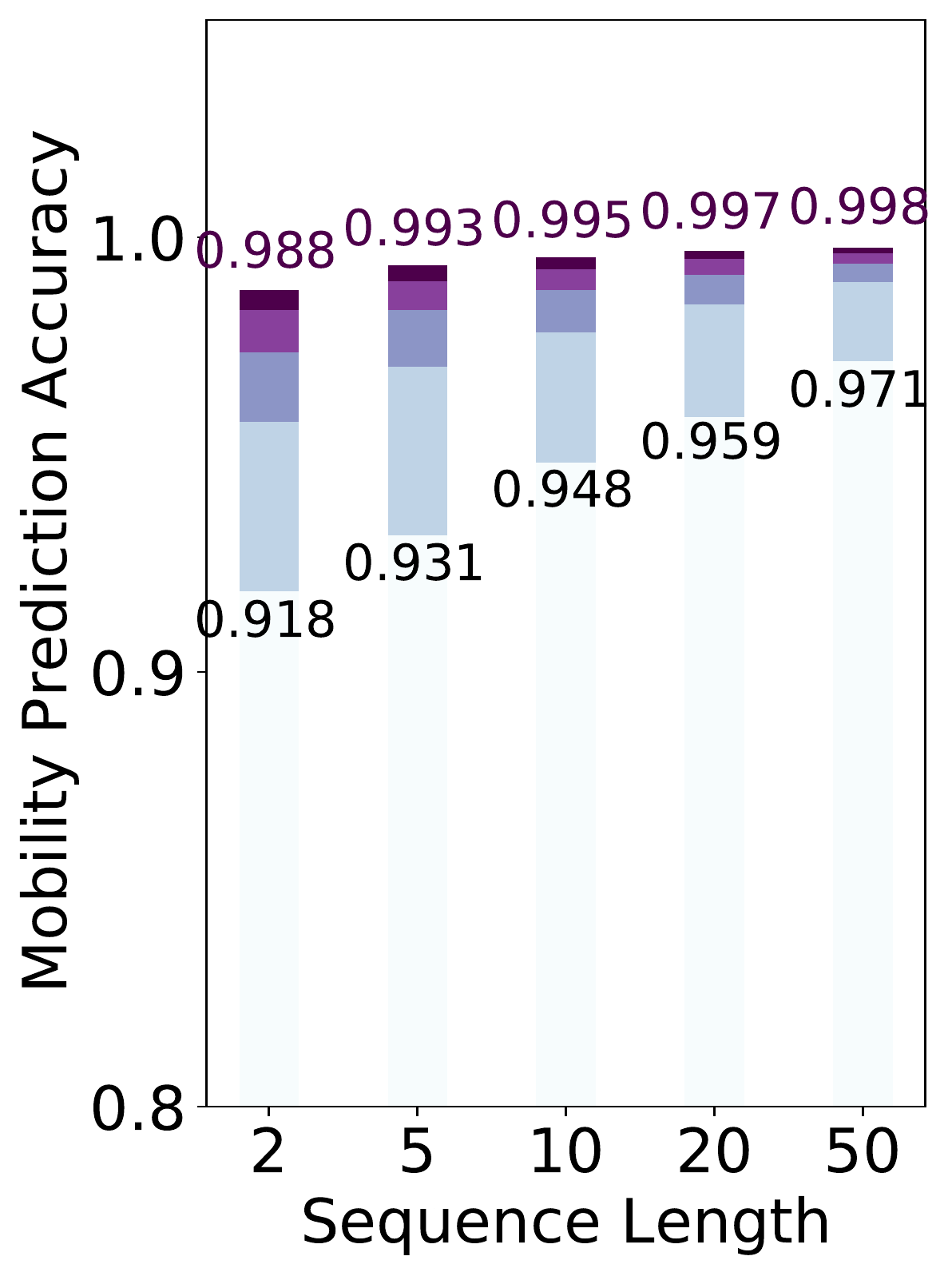}
         \caption{$U_D$ Performance on Priva'Mov}
         \label{fig:context_pri_prediction}
     \end{subfigure}
     \hfill
     \begin{subfigure}[b]{0.24\textwidth}
         \centering
         \includegraphics[width=\textwidth]{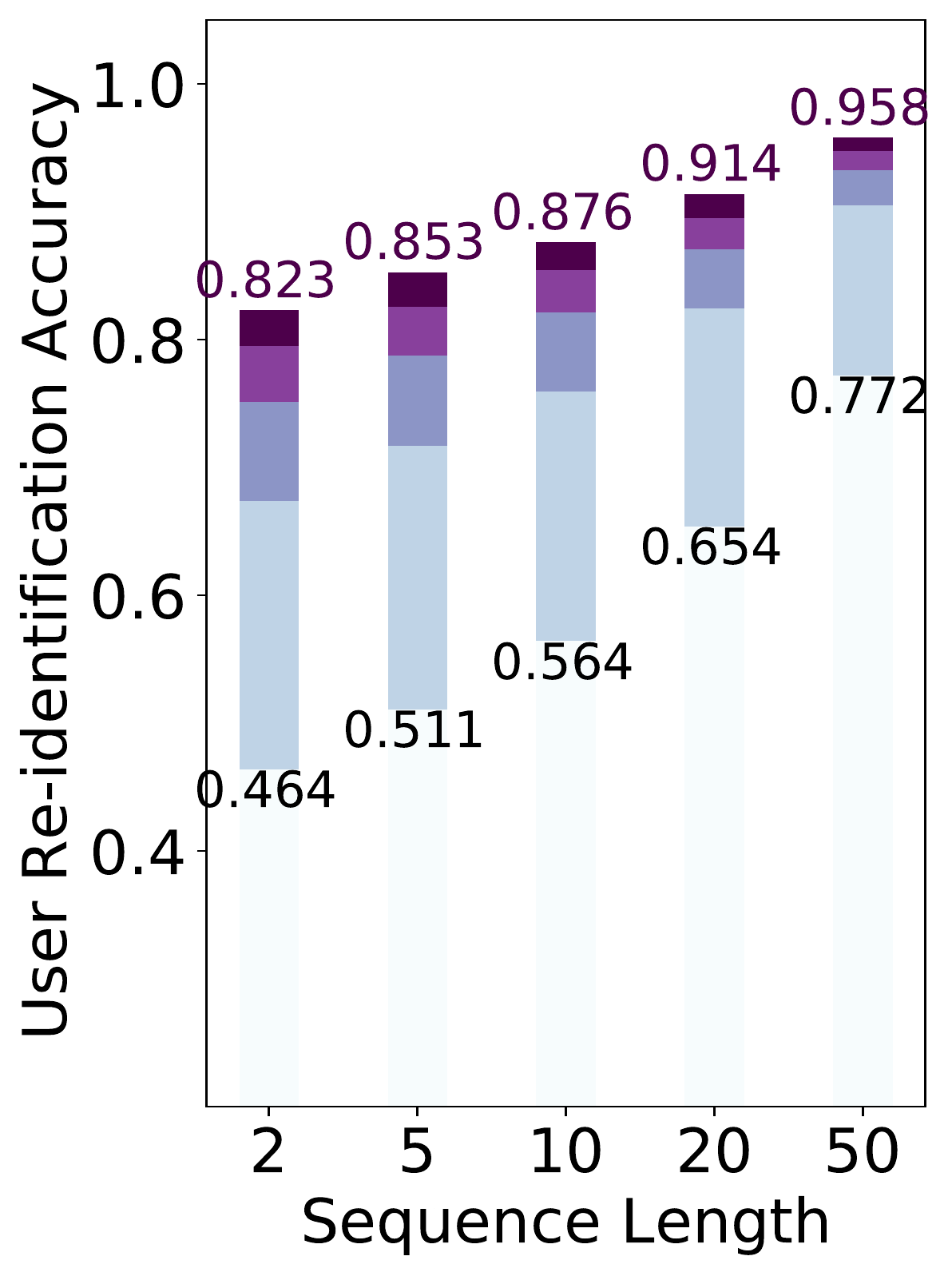}
         \caption{$P_D^2$ Performance on Priva'Mov}
         \label{fig:context_mdc_recognition2}
    \end{subfigure}
    \hfill
    \caption{Mobility prediction accuracy and user re-identification accuracy change with the trace sequence length (SL) in our proposed $U_D$ and $P_D^2$. The color bars indicate the accuracy from top-1 to top-5, the black texts indicate the top-1 accuracy and the purple texts indicate the top-5 accuracy. For instance, the top-1 mobility prediction accuracy on MDC with \textit{SL} = 2 is 0.473, and the top-5 one is 0.802.}
    %\Description{The effect of the context length}
    \label{fig:context}
\end{figure*}

\subsection{Privacy Guarantee Analysis: Effectiveness of Privacy Inference attacks}
\label{UPA}

In this section, we discuss the impact of Mo-PAE on the effectiveness of two privacy inference attacks (i.e., \textit{PI} and \textit{PII}), respectively.

\subsubsection{Effectiveness of Data Reconstruction Attacks - PI}

Table~\ref{tab:euc} shows the impact of the proposed mechanisms on the data reconstruction accuracy(\emph{PI}). The \emph{"Euc"} in the table follows the definition in Eq.\ref{eq:euc}. Overall, \emph{Model II} performs better than \emph{Model I} in limiting the accuracy of data reconstruction regardless of the value of weights. Take the result of GeoLife dataset as an example, \emph{Model II-i} achieves bigger distance than \emph{Model I} (i.e., 0.4343 > 0.0057), while it still gets better utility (i.e., -9.94\% > -17.9\%). Nevertheless, both \emph{Model I} and \emph{Model II} have effectively defended the data reconstruction attack (MDC: 0.0697 > 0.0017 > 0.0000; Priva'Mov: 0.0453 > 0.0009 > 0.0003; GeoLife: 0.4343 > 0.0057 > 0.0008; FourSquare: 0.7933 > 0.0069 > 0.0052), while only at the marginal cost of mobility utility (MDC: -12.55\%; Priva'Mov: -2.71\%; GeoLife: -9.94\%; FourSquare: -1.64\%). The data of the Optimal-IMs are in Table \ref{tab:compare}. We list four representative settings here to make a comprehensive comparison of PII and U. From setting i to iv, one can expect more original data features loss to result in a more significant utility loss. This trend is indeed the case with different weights' combinations. However, as the results show, especially for setting i, the privacy of the traces attains decent protection at the marginal cost of mobility utility.

\subsubsection{Effectiveness of User Re-identification Attacks - PII}
Figure~\ref{fig:utility-privacy} presents the impact of the Mo-PAE (\emph{Model II}) on the user re-identification accuracy(\emph{PII}). In this figure, we list five different settings, I to V ($\lambda_1 = 0.1$, over the range of $\lambda_2$ = \{0.5, 0.6, 0.7, 0.8, 0.9\}), respectively.
The \textit{Zero} line (i.e., y = 0\%) in each sub-figure is leveraged to indicate the original accuracy of the raw data (i.e., Optimal-IMs). The \emph{"Privacy Gain Rate"} (blue square line) shows the effectiveness of the Mo-PAE in defending the user re-identification attacks. That is, after applying \emph{Model II}, the decline range of effectiveness of user re-identification attacks. For instance, with the MDC dataset, in setting I, the effectiveness of user re-identification attacks declines as high as 80\%. In the same time, this high privacy protection is at the cost of nearly 55\% of utility (orange triangle line). Things are better in the setting V, where the Mo-PAE can get 60\% privacy protection only at the cost of less than 10\% utility. The x-axis shows five settings of the model, and the y-axis shows the trade-offs (i.e., \textit{trade-offs} = \textit{privacy gain} + \textit{utility loss}). The orange area represents the utility loss while the light-green area represents the privacy gain when compared with Optimal-IMs. The dark-green area represents the trade-offs between utility and privacy budgets. 

In summary, these trade-offs are all positive in different model settings on four different datasets. 
The performance on the GeoLife data is the best, while less than 20\% utility loss but more than 50\% privacy gains. The performance on MDC and FourSquare also show the promising privacy-utility trade-offs, especially for setting \textit{V} on the FourSquare dataset, both the utility and privacy increase.
The uniqueness of human mobility trajectories is high, and these trajectories are likely to be re-identified even with a few location data points~\cite{de2013unique}. Our results emphasize that the concern of user re-identification risk could be alleviated effectively with our proposed model.

In real applications, the trade-off of Mo-PAE is achieved continuously over time. New trajectories will be encoded with the pre-trained encoder to attain respective feature representation and utilized by SP for following task-oriented scenarios (no need to retrain). The pre-trained encoder and discriminators are assumed to be updated offline within a fixed duration for best performance purposes. 
Additionally, while the architecture focuses on specific application scenarios (i.e., mobility prediction), it could generally be applicable to different task-oriented scenarios. 

%\subsubsection{Generalization Performance of Mo-PAE}

\section{Discussions}
\label{sec:discussions}

In this section, we further discuss the impact of the temporal granularity of traces, the varying sequence length and weights on the composition units on the Mo-PAE performance.

\subsection{Impact of Temporal Granularity}

The timestamp is one of the essential components of the mobility trace, and different choices on the temporal granularity affect the final performance of any dataset. Figure~\ref{fig:temporal} shows the impact of the varying temporal granularity on the proposed architecture. We present the top-1, top-5, and top-10 accuracies for both utility and privacy dimensions. For instance, when temporal granularity is 10-min, it indicates a location record \textit{r} is taken every 10 minutes from the raw data. When using more coarser temporal granularity, the number of points of interest decreases so does the difficulty of mobility prediction. However, the uniqueness of the trajectory decreases due to ignoring many of the unique locations of each user, resulting in lower privacy. To summarize from Figure~\ref{fig:temporal}, the impact of temporal granularity on the Priva'Mov is minimal. In terms of utility (mobility prediction), Priva'Mov is the only dataset for which accuracy decreases with increasing temporal granularity. This subtle decline emphasizes the trajectory features only have a small change when varying granularity, in line with the university students' mobility.

\subsection{Impact of Varying Sequence Lengths}
\label{section:impact sl}
The performance of the utility discriminator $U_D$ (MPU) and the privacy discriminator $P_D^2$ (URU) exert a significant impact on the overall performance of the proposed Mo-PAE. The trace length is the most critical factor affecting these units' performance. We use two representative datasets (i.e., MDC and Priva'Mov) to present the impact of the varying sequence length on both discriminators. 

As shown in Figure~\ref{fig:context}, by changing the lengths of trace sequence $SL$ from 1 ($SL=1$) to 50 ($SL=50$), we observe that $SL$ has a significant impact on different tasks' accuracy (i.e., mobility prediction accuracy for $U_D$ and user re-identification accuracy for $P_D^2$) of two different datasets. Overall, the impact in the MDC dataset is much higher than in the Priva'Mov dataset. Comparing the Figure~\ref{fig:context_mdc_prediction} and Figure~\ref{fig:context_pri_prediction}, there is a much sharper increase on the MDC dataset. More specifically, when the sequence length is increased from 2 to 20, the top-1 mobility prediction accuracy on MDC increases from 0.473 to 0.978 (i.e., +50.5\%), while accuracy on Priva'Mov increases from 0.918 to 0.959 (i.e., only +4.1\%). Similarly, more rapid growth appears in the user re-identification accuracy on MDC, which is +68.0\%, while the increase for Priva'Mov is only +30.8\%. We conclude that the mobility predictability and user re-identification accuracy of a dataset might have a special link. The mobility predictability of Priva'Mov is very high, almost higher than 90\%, but the user re-identification accuracy is always lower than 80\%, which also means the uniqueness of trajectories in this dataset is low. This low uniqueness suggests that the users in this dataset might share similar daily routes, which is reasonable, as we know these trajectories are collected from students at the same university. For the MDC dataset, when $SL=10$, the user re-identification accuracy is relatively high, indicating that the locations are more sparse in this dataset. However, the mobility predictability here is also high, which emphasizing that this sparseness does not affect the predictability. These phenomena indicate that the deep training of MPU and URU might share similar extracted features, while our proposed architecture attempts to extract features more suitable for mobility predictability but less suitable for user re-identification.

We note that the varying trace sequence length not only exerts impacts on the model performance but also has a significant influence on the computation time. For instance, the computation time at $SL=50$ costs six times as much as that at $SL=5$. The computation time also varies between datasets. Hence, an appropriate choice of trajectory sequence length can avoid time-consuming computation and achieve expected task inference accuracy. In our work, we place greater focus on the trace sequence lengths ranging from 5 to 10, which exhibits great performance in both the $U_D$ and $P_D$ while also keeping a low computation time.

\begin{figure}
     \centering
     \begin{subfigure}[b]{0.48\textwidth}
         \centering
         \includegraphics[width=\textwidth]{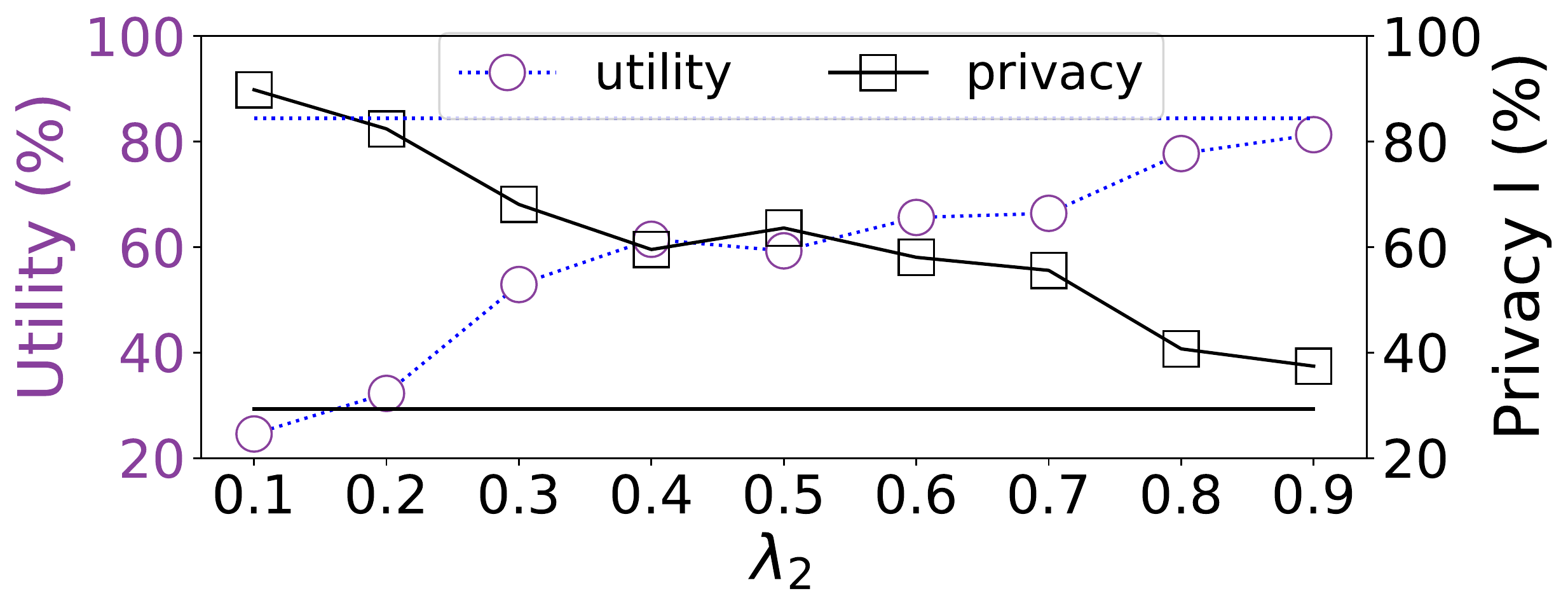}
         \caption{$\lambda_1 = 1-\lambda_2,~\lambda_3 = 0$}
         \label{fig:lambda3_zero_privacy}
     \end{subfigure}
     \begin{subfigure}[b]{0.48\textwidth}
         \centering
         \includegraphics[width=\textwidth]{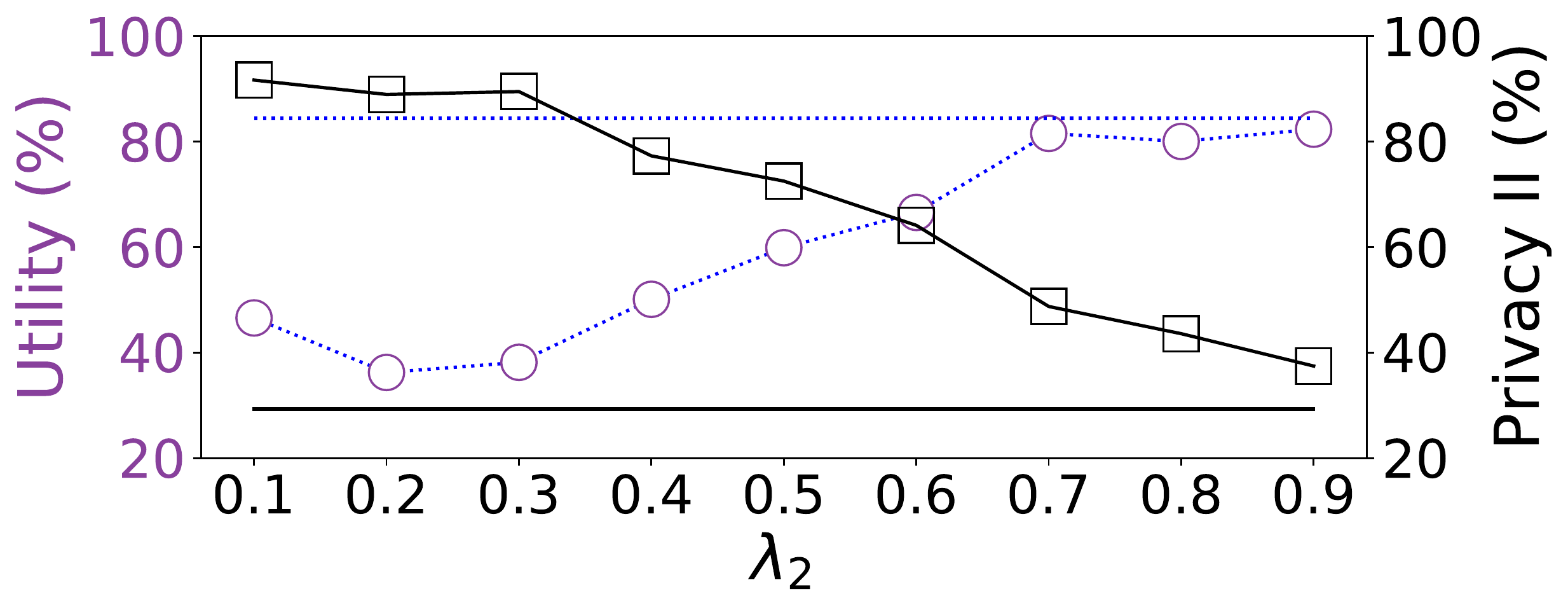}
         \caption{$\lambda_3 = 1-\lambda_2,~\lambda_1 = 0$}
         \label{fig:lambda1_zero_privacy}
    \end{subfigure}
    \caption{Varying weights can tune the privacy-utility trade-offs. The primary y-axis (dashed line) represents utility, the secondary y-axis (solid line) represents privacy. The x-axis represents the value of the target $\lambda$. }
    %\Description{privacy and accuracy}
    \label{fig:lagranian}
\end{figure}

\subsection{Impact of Varying Weights}% in Loss Function}
As we discussed in Section~\ref{section: module overview}, the \textit{sum loss function} $L_{sum}$ of \emph{Model II} is a linear combination of $L_R$, $L_U$, and $L_P$ with different weights (i.e., Lagrange multipliers). We evaluate the influence of different weights' combinations ($\lambda_1$, $\lambda_2$, and $\lambda_3$) on the \emph{Model II}, as the results shown in Figure~\ref{fig:lagranian}.

We compare the overall model performance in $U_D$ and $P_D^1$ by fixing the $\lambda_3 = 0$, and vary the other two multipliers by subjecting to $\lambda_1 = 1- \lambda_2$, as shown in Figure~\ref{fig:lambda3_zero_privacy}. Figure~\ref{fig:lambda1_zero_privacy} illustrates the effect between $U_D$ and $P_D^2$ by setting the $\lambda_1 = 0$. We could observe in both settings that the utility increases with a larger $\lambda_2$, which means when the MPU is given more weight in the Mo-PAE model, it would exert a positive impact on the data utility. We conclude that the privacy-utility trade-offs could be tuned by varying these weights; the results in the Figure~\ref{fig:lagranian} also verify the effectiveness of our adversarial architecture. We note that the balance of three units is far more complicated than the balance of two. From the extensive experiment we conducted, initialing $\lambda_1=0.1$, $\lambda_2=0.6$, $\lambda_3=0.3$ can guide the model achieve the tradeoff most efficiently. 
However, as the experiment results show, there is no dataset independent privacy interpretation for $\lambda_1$, $\lambda_2$ and $\lambda_3$, and we leave a more efficient approach using reinforcement learning to initializing these hyperparameters for different datasets in future work.
%However, this work primarily focuses on the applicability of applying adversarial learning and representation learning to attain the privacy-utility tradeoff. We understand the best combination mentioned in the paper might not lead to the global optimum, and new approach to parametric them using reinforcement learning will be part of our future work. We will add this in Discussion.

\section{Conclusion}
\label{Conl}

In this paper, we presented a privacy-preserving architecture Mo-PAE based on adversarial networks. Our model considers three different optimization objectives and searches for the optimum trade-off for utility and privacy of a given dataset. We reported an extensive analysis of our model performances and the impact of its hyperparameters using four real-world mobility datasets. The weights $\lambda_1,\ \lambda_2,$ and $\lambda_3$ bring more flexibility to our framework, enabling it to satisfy different scenarios' requirements according to the relative importance of utility requirements and privacy budgets. We evaluated our framework on four datasets and benchmarked our results against an LSTM-GAN approach and a DP mechanism. The comparisons indicate the superiority of the proposed framework and the efficiency of the proposed privacy-preserving feature extractor $Enc_L$. Expanding this work, we will consider other utility functions for our models, such as community detection based on unsupervised clustering methods or deep embedded clustering methods. In future work, we will leverage automated search techniques, such as deep deterministic policy gradient algorithm and reinforcement learning, for efficiency in searching for the optimal weight combinations.

\section*{Acknowledgment}
We wish to acknowledge the constructive suggestions from PETS reviewers. This work was partially supported by EPSRC Open Plus Fellowship EP/W005271/1.

\bibliographystyle{ACM-Reference-Format}
\bibliography{main}

%%% -*-BibTeX-*-
%%% Do NOT edit. File created by BibTeX with style
%%% ACM-Reference-Format-Journals [18-Jan-2012].

\begin{thebibliography}{62}

%%% ====================================================================
%%% NOTE TO THE USER: you can override these defaults by providing
%%% customized versions of any of these macros before the \bibliography
%%% command.  Each of them MUST provide its own final punctuation,
%%% except for \shownote{}, \showDOI{}, and \showURL{}.  The latter two
%%% do not use final punctuation, in order to avoid confusing it with
%%% the Web address.
%%%
%%% To suppress output of a particular field, define its macro to expand
%%% to an empty string, or better, \unskip, like this:
%%%
%%% \newcommand{\showDOI}[1]{\unskip}   % LaTeX syntax
%%%
%%% \def \showDOI #1{\unskip}           % plain TeX syntax
%%%
%%% ====================================================================

\ifx \showCODEN    \undefined \def \showCODEN     #1{\unskip}     \fi
\ifx \showDOI      \undefined \def \showDOI       #1{#1}\fi
\ifx \showISBNx    \undefined \def \showISBNx     #1{\unskip}     \fi
\ifx \showISBNxiii \undefined \def \showISBNxiii  #1{\unskip}     \fi
\ifx \showISSN     \undefined \def \showISSN      #1{\unskip}     \fi
\ifx \showLCCN     \undefined \def \showLCCN      #1{\unskip}     \fi
\ifx \shownote     \undefined \def \shownote      #1{#1}          \fi
\ifx \showarticletitle \undefined \def \showarticletitle #1{#1}   \fi
\ifx \showURL      \undefined \def \showURL       {\relax}        \fi
% The following commands are used for tagged output and should be
% invisible to TeX
\providecommand\bibfield[2]{#2}
\providecommand\bibinfo[2]{#2}
\providecommand\natexlab[1]{#1}
\providecommand\showeprint[2][]{arXiv:#2}

\bibitem[{Aktay} et~al\mbox{.}(2020)]%
        {aktay2020google}
\bibfield{author}{\bibinfo{person}{Ahmet {Aktay}}, \bibinfo{person}{Shailesh
  {Bavadekar}}, \bibinfo{person}{Gwen {Cossoul}}, \bibinfo{person}{John
  {Davis}}, \bibinfo{person}{Damien {Desfontaines}}, \bibinfo{person}{Alex
  {Fabrikant}}, \bibinfo{person}{Evgeniy {Gabrilovich}},
  \bibinfo{person}{Krishna {Gadepalli}}, \bibinfo{person}{Bryant {Gipson}},
  \bibinfo{person}{Miguel {Guevara}}, \bibinfo{person}{Chaitanya {Kamath}},
  \bibinfo{person}{Mansi {Kansal}}, \bibinfo{person}{Ali {Lange}},
  \bibinfo{person}{Chinmoy {Mandayam}}, \bibinfo{person}{Andrew {Oplinger}},
  \bibinfo{person}{Christopher {Pluntke}}, \bibinfo{person}{Thomas {Roessler}},
  \bibinfo{person}{Arran {Schlosberg}}, \bibinfo{person}{Tomer {Shekel}},
  \bibinfo{person}{Swapnil {Vispute}}, \bibinfo{person}{Mia {Vu}},
  \bibinfo{person}{Gregory {Wellenius}}, \bibinfo{person}{Brian {Williams}},
  {and} \bibinfo{person}{Royce~J {Wilson}}.} \bibinfo{year}{2020}\natexlab{}.
\newblock \bibinfo{title}{{Google {COVID-19} Community Mobility Reports:
  Anonymization Process Description (version 1.1)}}.
\newblock , \bibinfo{numpages}{arXiv:2004.04145}~pages.
\newblock
\urldef\tempurl%
\url{https://doi.org/10.48550/ARXIV.2004.04145}
\showDOI{\tempurl}


\bibitem[Andr\'{e}s et~al\mbox{.}(2013)]%
        {andres2013geo}
\bibfield{author}{\bibinfo{person}{Miguel~E. Andr\'{e}s},
  \bibinfo{person}{Nicol\'{a}s~E. Bordenabe}, \bibinfo{person}{Konstantinos
  Chatzikokolakis}, {and} \bibinfo{person}{Catuscia Palamidessi}.}
  \bibinfo{year}{2013}\natexlab{}.
\newblock \showarticletitle{Geo-Indistinguishability: Differential Privacy for
  Location-Based Systems}. In \bibinfo{booktitle}{\emph{Proceedings of the 2013
  ACM SIGSAC Conference on Computer \& Communications Security}} (Berlin,
  Germany) \emph{(\bibinfo{series}{CCS '13})}. \bibinfo{publisher}{Association
  for Computing Machinery}, \bibinfo{address}{New York, NY, USA},
  \bibinfo{pages}{901–914}.
\newblock
\showISBNx{9781450324779}
\urldef\tempurl%
\url{https://doi.org/10.1145/2508859.2516735}
\showDOI{\tempurl}


\bibitem[Ball(1960)]%
        {ball1960short}
\bibfield{author}{\bibinfo{person}{Walter William~Rouse Ball}.}
  \bibinfo{year}{1960}\natexlab{}.
\newblock \bibinfo{booktitle}{\emph{A short account of the history of
  mathematics}}.
\newblock \bibinfo{publisher}{Courier Corporation}, \bibinfo{address}{New York,
  NY, USA}.
\newblock


\bibitem[Beavis and Dobbs(1990)]%
        {beavis1990optimisation}
\bibfield{author}{\bibinfo{person}{Brian Beavis} {and} \bibinfo{person}{Ian
  Dobbs}.} \bibinfo{year}{1990}\natexlab{}.
\newblock \bibinfo{booktitle}{\emph{Optimisation and Stability Theory for
  Economic Analysis}}.
\newblock \bibinfo{publisher}{Cambridge University Press},
  \bibinfo{address}{Cambridge}.
\newblock
\urldef\tempurl%
\url{https://doi.org/10.1017/CBO9780511559402}
\showDOI{\tempurl}


\bibitem[Ben~Mokhtar et~al\mbox{.}(2017)]%
        {mokhtar2017priva}
\bibfield{author}{\bibinfo{person}{Sonia Ben~Mokhtar}, \bibinfo{person}{Antoine
  Boutet}, \bibinfo{person}{Louafi Bouzouina}, \bibinfo{person}{Patrick
  Bonnel}, \bibinfo{person}{Olivier Brette}, \bibinfo{person}{Lionel Brunie},
  \bibinfo{person}{Mathieu Cunche}, \bibinfo{person}{Stephane D~'Alu},
  \bibinfo{person}{Vincent Primault}, \bibinfo{person}{Patrice Raveneau},
  \bibinfo{person}{Herve Rivano}, {and} \bibinfo{person}{Razvan Stanica}.}
  \bibinfo{year}{2017}\natexlab{}.
\newblock \showarticletitle{{PRIVA'MOV: Analysing Human Mobility Through
  Multi-Sensor Datasets}}. In \bibinfo{booktitle}{\emph{{NetMob 2017}}}.
  \bibinfo{publisher}{HAL-Inria}, \bibinfo{address}{Milan, Italy}.
\newblock
\urldef\tempurl%
\url{https://hal.inria.fr/hal-01578557}
\showURL{%
\tempurl}


\bibitem[Beresford and Stajano(2003)]%
        {beresford2003location}
\bibfield{author}{\bibinfo{person}{Alastair~R Beresford} {and}
  \bibinfo{person}{Frank Stajano}.} \bibinfo{year}{2003}\natexlab{}.
\newblock \showarticletitle{Location privacy in pervasive computing}.
\newblock \bibinfo{journal}{\emph{IEEE Pervasive computing}}
  \bibinfo{volume}{2}, \bibinfo{number}{1} (\bibinfo{year}{2003}),
  \bibinfo{pages}{46--55}.
\newblock


\bibitem[Bindschaedler and Shokri(2016)]%
        {bindschaedler2016synthesizing}
\bibfield{author}{\bibinfo{person}{Vincent Bindschaedler} {and}
  \bibinfo{person}{Reza Shokri}.} \bibinfo{year}{2016}\natexlab{}.
\newblock \showarticletitle{Synthesizing Plausible Privacy-Preserving Location
  Traces}. In \bibinfo{booktitle}{\emph{2016 IEEE Symposium on Security and
  Privacy (SP)}}. \bibinfo{publisher}{{IEEE}}, \bibinfo{address}{San Jose, CA,
  USA}, \bibinfo{pages}{546--563}.
\newblock
\urldef\tempurl%
\url{https://doi.org/10.1109/SP.2016.39}
\showDOI{\tempurl}


\bibitem[Black(2020)]%
        {black1998dictionary}
\bibfield{author}{\bibinfo{person}{Paul~E Black}.}
  \bibinfo{year}{2020}\natexlab{}.
\newblock \bibinfo{booktitle}{\emph{{DADS:The On-Line Dictionary of Algorithms
  and Data Structures}}}.
\newblock \bibinfo{type}{{T}echnical {R}eport}. \bibinfo{institution}{National
  Institute of Standards and Technology}.
\newblock
\urldef\tempurl%
\url{https://doi.org/10.6028/nist.ir.8318}
\showDOI{\tempurl}


\bibitem[Chatzikokolakis et~al\mbox{.}(2015)]%
        {chatzikokolakis2015geo}
\bibfield{author}{\bibinfo{person}{Konstantinos Chatzikokolakis},
  \bibinfo{person}{Catuscia Palamidessi}, {and} \bibinfo{person}{Marco
  Stronati}.} \bibinfo{year}{2015}\natexlab{}.
\newblock \showarticletitle{Geo-indistinguishability: A Principled Approach to
  Location Privacy}. In \bibinfo{booktitle}{\emph{Distributed Computing and
  Internet Technology}}, \bibfield{editor}{\bibinfo{person}{Raja Natarajan},
  \bibinfo{person}{Gautam Barua}, {and} \bibinfo{person}{Manas~Ranjan Patra}}
  (Eds.). \bibinfo{publisher}{Springer International Publishing},
  \bibinfo{address}{Cham}, \bibinfo{pages}{49--72}.
\newblock
\showISBNx{978-3-319-14977-6}


\bibitem[Choi et~al\mbox{.}(2021)]%
        {choi2021trajgail}
\bibfield{author}{\bibinfo{person}{Seongjin Choi}, \bibinfo{person}{Jiwon Kim},
  {and} \bibinfo{person}{Hwasoo Yeo}.} \bibinfo{year}{2021}\natexlab{}.
\newblock \showarticletitle{TrajGAIL: Generating urban vehicle trajectories
  using generative adversarial imitation learning}.
\newblock \bibinfo{journal}{\emph{Transportation Research Part C: Emerging
  Technologies}}  \bibinfo{volume}{128} (\bibinfo{year}{2021}),
  \bibinfo{pages}{103091}.
\newblock


\bibitem[Clifton and Tassa(2013)]%
        {clifton2013syntactic}
\bibfield{author}{\bibinfo{person}{Chris Clifton} {and} \bibinfo{person}{Tamir
  Tassa}.} \bibinfo{year}{2013}\natexlab{}.
\newblock \showarticletitle{On syntactic anonymity and differential privacy}.
  In \bibinfo{booktitle}{\emph{2013 IEEE 29th International Conference on Data
  Engineering Workshops (ICDEW)}}. \bibinfo{publisher}{{IEEE}},
  \bibinfo{address}{Brisbane, QLD, Australia}, \bibinfo{pages}{88--93}.
\newblock


\bibitem[Cunningham et~al\mbox{.}(2021)]%
        {cunningham2021privacy}
\bibfield{author}{\bibinfo{person}{Teddy Cunningham}, \bibinfo{person}{Graham
  Cormode}, {and} \bibinfo{person}{Hakan Ferhatosmanoglu}.}
  \bibinfo{year}{2021}\natexlab{}.
\newblock \bibinfo{booktitle}{\emph{Privacy-Preserving Synthetic Location Data
  in the Real World}}.
\newblock \bibinfo{publisher}{Association for Computing Machinery},
  \bibinfo{address}{New York, NY, USA}, \bibinfo{pages}{23–33}.
\newblock
\showISBNx{9781450384254}
\urldef\tempurl%
\url{https://doi.org/10.1145/3469830.3470893}
\showURL{%
\tempurl}


\bibitem[de~Mattos et~al\mbox{.}(2019)]%
        {de2019give}
\bibfield{author}{\bibinfo{person}{Ekler~P de Mattos},
  \bibinfo{person}{Augusto~CSA Domingues}, {and} \bibinfo{person}{Antonio~AF
  Loureiro}.} \bibinfo{year}{2019}\natexlab{}.
\newblock \showarticletitle{Give me two points and i'll tell you who you are}.
  In \bibinfo{booktitle}{\emph{2019 IEEE Intelligent Vehicles Symposium (IV)}}.
  \bibinfo{publisher}{{IEEE}}, \bibinfo{address}{Paris, France},
  \bibinfo{pages}{1081--1087}.
\newblock


\bibitem[De~Montjoye et~al\mbox{.}(2013)]%
        {de2013unique}
\bibfield{author}{\bibinfo{person}{Yves-Alexandre De~Montjoye},
  \bibinfo{person}{C{\'e}sar~A Hidalgo}, \bibinfo{person}{Michel Verleysen},
  {and} \bibinfo{person}{Vincent~D Blondel}.} \bibinfo{year}{2013}\natexlab{}.
\newblock \showarticletitle{Unique in the crowd: The privacy bounds of human
  mobility}.
\newblock \bibinfo{journal}{\emph{Scientific reports}}  \bibinfo{volume}{3}
  (\bibinfo{year}{2013}), \bibinfo{pages}{1376}.
\newblock


\bibitem[Dwork(2008)]%
        {dwork2008differential}
\bibfield{author}{\bibinfo{person}{Cynthia Dwork}.}
  \bibinfo{year}{2008}\natexlab{}.
\newblock \showarticletitle{Differential Privacy: A Survey of Results}. In
  \bibinfo{booktitle}{\emph{Theory and Applications of Models of Computation}},
  \bibfield{editor}{\bibinfo{person}{Manindra Agrawal},
  \bibinfo{person}{Dingzhu Du}, \bibinfo{person}{Zhenhua Duan}, {and}
  \bibinfo{person}{Angsheng Li}} (Eds.). \bibinfo{publisher}{Springer Berlin
  Heidelberg}, \bibinfo{address}{Berlin, Heidelberg}, \bibinfo{pages}{1--19}.
\newblock
\showISBNx{978-3-540-79228-4}


\bibitem[Dwork et~al\mbox{.}(2014)]%
        {dwork2014algorithmic}
\bibfield{author}{\bibinfo{person}{Cynthia Dwork}, \bibinfo{person}{Aaron
  Roth}, {et~al\mbox{.}}} \bibinfo{year}{2014}\natexlab{}.
\newblock \showarticletitle{The algorithmic foundations of differential
  privacy.}
\newblock \bibinfo{journal}{\emph{Found. Trends Theor. Comput. Sci.}}
  \bibinfo{volume}{9}, \bibinfo{number}{3-4} (\bibinfo{year}{2014}),
  \bibinfo{pages}{211--407}.
\newblock


\bibitem[Erdemir et~al\mbox{.}(2021)]%
        {erdemir_privacy-aware_2020}
\bibfield{author}{\bibinfo{person}{Ecenaz Erdemir}, \bibinfo{person}{Pier~Luigi
  Dragotti}, {and} \bibinfo{person}{Deniz Gündüz}.}
  \bibinfo{year}{2021}\natexlab{}.
\newblock \showarticletitle{Privacy-Aware Time-Series Data Sharing With Deep
  Reinforcement Learning}.
\newblock \bibinfo{journal}{\emph{IEEE Transactions on Information Forensics
  and Security}}  \bibinfo{volume}{16} (\bibinfo{year}{2021}),
  \bibinfo{pages}{389--401}.
\newblock
\urldef\tempurl%
\url{https://doi.org/10.1109/TIFS.2020.3013200}
\showDOI{\tempurl}


\bibitem[Errounda and Liu(2019)]%
        {errounda2019analysis}
\bibfield{author}{\bibinfo{person}{Fatima~Zahra Errounda} {and}
  \bibinfo{person}{Yan Liu}.} \bibinfo{year}{2019}\natexlab{}.
\newblock \showarticletitle{An analysis of differential privacy research in
  location data}. In \bibinfo{booktitle}{\emph{2019 IEEE 5th Intl Conference on
  Big Data Security on Cloud (BigDataSecurity), IEEE Intl Conference on High
  Performance and Smart Computing,(HPSC) and IEEE Intl Conference on
  Intelligent Data and Security (IDS)}}. \bibinfo{publisher}{{IEEE}},
  \bibinfo{address}{Washington, DC, USA}, \bibinfo{pages}{53--60}.
\newblock


\bibitem[Feng et~al\mbox{.}(2020)]%
        {feng2020pmf}
\bibfield{author}{\bibinfo{person}{Jie Feng}, \bibinfo{person}{Can Rong},
  \bibinfo{person}{Funing Sun}, \bibinfo{person}{Diansheng Guo}, {and}
  \bibinfo{person}{Yong Li}.} \bibinfo{year}{2020}\natexlab{}.
\newblock \showarticletitle{PMF: A privacy-preserving human mobility prediction
  framework via federated learning}.
\newblock \bibinfo{journal}{\emph{Proceedings of the ACM on Interactive,
  Mobile, Wearable and Ubiquitous Technologies}} \bibinfo{volume}{4},
  \bibinfo{number}{1} (\bibinfo{year}{2020}), \bibinfo{pages}{1--21}.
\newblock


\bibitem[Ferreira et~al\mbox{.}(2020)]%
        {ferreira_deep_2020}
\bibfield{author}{\bibinfo{person}{Danielle~L. Ferreira},
  \bibinfo{person}{Bruno A.~A. Nunes}, \bibinfo{person}{Carlos Alberto~V.
  Campos}, {and} \bibinfo{person}{Katia Obraczka}.}
  \bibinfo{year}{2020}\natexlab{}.
\newblock \showarticletitle{A {Deep} {Learning} {Approach} for {Identifying}
  {User} {Communities} {Based} on {Geographical} {Preferences} and {Its}
  {Applications} to {Urban} and {Environmental} {Planning}}.
\newblock \bibinfo{journal}{\emph{ACM Transactions on Spatial Algorithms and
  Systems}} \bibinfo{volume}{6}, \bibinfo{number}{3} (\bibinfo{date}{May}
  \bibinfo{year}{2020}), \bibinfo{pages}{1--24}.
\newblock
\showISSN{2374-0353, 2374-0361}
\urldef\tempurl%
\url{https://doi.org/10.1145/3380970}
\showDOI{\tempurl}


\bibitem[Gedik and Liu(2005)]%
        {gedik2005location}
\bibfield{author}{\bibinfo{person}{Bugra Gedik} {and} \bibinfo{person}{Ling
  Liu}.} \bibinfo{year}{2005}\natexlab{}.
\newblock \showarticletitle{Location privacy in mobile systems: A personalized
  anonymization model}. In \bibinfo{booktitle}{\emph{25th IEEE International
  Conference on Distributed Computing Systems (ICDCS'05)}}.
  \bibinfo{publisher}{{IEEE}}, \bibinfo{address}{Columbus, OH, USA},
  \bibinfo{pages}{620--629}.
\newblock


\bibitem[Gedik and Liu(2007)]%
        {gedik2007protecting}
\bibfield{author}{\bibinfo{person}{Bugra Gedik} {and} \bibinfo{person}{Ling
  Liu}.} \bibinfo{year}{2007}\natexlab{}.
\newblock \showarticletitle{Protecting location privacy with personalized
  k-anonymity: Architecture and algorithms}.
\newblock \bibinfo{journal}{\emph{IEEE Transactions on Mobile Computing}}
  \bibinfo{volume}{7}, \bibinfo{number}{1} (\bibinfo{year}{2007}),
  \bibinfo{pages}{1--18}.
\newblock


\bibitem[Gomes et~al\mbox{.}(2013)]%
        {gomes2013will}
\bibfield{author}{\bibinfo{person}{Jo{\~a}o~B{\'a}rtolo Gomes},
  \bibinfo{person}{Clifton Phua}, {and} \bibinfo{person}{Shonali
  Krishnaswamy}.} \bibinfo{year}{2013}\natexlab{}.
\newblock \showarticletitle{Where Will You Go? Mobile Data Mining for Next
  Place Prediction}. In \bibinfo{booktitle}{\emph{Data Warehousing and
  Knowledge Discovery}}, \bibfield{editor}{\bibinfo{person}{Ladjel Bellatreche}
  {and} \bibinfo{person}{Mukesh~K. Mohania}} (Eds.).
  \bibinfo{publisher}{Springer Berlin Heidelberg}, \bibinfo{address}{Berlin,
  Heidelberg}, \bibinfo{pages}{146--158}.
\newblock
\showISBNx{978-3-642-40131-2}


\bibitem[Gonzalez et~al\mbox{.}(2008)]%
        {gonzalez2008understanding}
\bibfield{author}{\bibinfo{person}{Marta~C Gonzalez}, \bibinfo{person}{Cesar~A
  Hidalgo}, {and} \bibinfo{person}{Albert-Laszlo Barabasi}.}
  \bibinfo{year}{2008}\natexlab{}.
\newblock \showarticletitle{Understanding individual human mobility patterns}.
\newblock \bibinfo{journal}{\emph{nature}} \bibinfo{volume}{453},
  \bibinfo{number}{7196} (\bibinfo{year}{2008}), \bibinfo{pages}{779--782}.
\newblock


\bibitem[Goodfellow et~al\mbox{.}(2014)]%
        {goodfellow2014generative}
\bibfield{author}{\bibinfo{person}{Ian Goodfellow}, \bibinfo{person}{Jean
  Pouget-Abadie}, \bibinfo{person}{Mehdi Mirza}, \bibinfo{person}{Bing Xu},
  \bibinfo{person}{David Warde-Farley}, \bibinfo{person}{Sherjil Ozair},
  \bibinfo{person}{Aaron Courville}, {and} \bibinfo{person}{Yoshua Bengio}.}
  \bibinfo{year}{2014}\natexlab{}.
\newblock \showarticletitle{Generative Adversarial Nets}. In
  \bibinfo{booktitle}{\emph{Advances in Neural Information Processing
  Systems}}, \bibfield{editor}{\bibinfo{person}{Z.~Ghahramani},
  \bibinfo{person}{M.~Welling}, \bibinfo{person}{C.~Cortes},
  \bibinfo{person}{N.~Lawrence}, {and} \bibinfo{person}{K.Q. Weinberger}}
  (Eds.), Vol.~\bibinfo{volume}{27}. \bibinfo{publisher}{Curran Associates,
  Inc.}, \bibinfo{address}{Montreal, Canada}.
\newblock
\urldef\tempurl%
\url{https://proceedings.neurips.cc/paper/2014/file/5ca3e9b122f61f8f06494c97b1afccf3-Paper.pdf}
\showURL{%
\tempurl}


\bibitem[Gursoy et~al\mbox{.}(2018)]%
        {gursoy2018utility}
\bibfield{author}{\bibinfo{person}{Mehmet~Emre Gursoy}, \bibinfo{person}{Ling
  Liu}, \bibinfo{person}{Stacey Truex}, \bibinfo{person}{Lei Yu}, {and}
  \bibinfo{person}{Wenqi Wei}.} \bibinfo{year}{2018}\natexlab{}.
\newblock \showarticletitle{Utility-Aware Synthesis of Differentially Private
  and Attack-Resilient Location Traces}. In
  \bibinfo{booktitle}{\emph{Proceedings of the 2018 ACM SIGSAC Conference on
  Computer and Communications Security}} (Toronto, Canada)
  \emph{(\bibinfo{series}{CCS '18})}. \bibinfo{publisher}{Association for
  Computing Machinery}, \bibinfo{address}{New York, NY, USA},
  \bibinfo{pages}{196–211}.
\newblock
\showISBNx{9781450356930}
\urldef\tempurl%
\url{https://doi.org/10.1145/3243734.3243741}
\showDOI{\tempurl}


\bibitem[Gursoy et~al\mbox{.}(2020)]%
        {gursoy2020utility}
\bibfield{author}{\bibinfo{person}{M~Emre Gursoy}, \bibinfo{person}{Vivekanand
  Rajasekar}, {and} \bibinfo{person}{Ling Liu}.}
  \bibinfo{year}{2020}\natexlab{}.
\newblock \showarticletitle{Utility-Optimized Synthesis of Differentially
  Private Location Traces}. In \bibinfo{booktitle}{\emph{2020 Second IEEE
  International Conference on Trust, Privacy and Security in Intelligent
  Systems and Applications (TPS-ISA)}}. \bibinfo{publisher}{{IEEE}},
  \bibinfo{address}{Atlanta, GA, USA}, \bibinfo{pages}{30--39}.
\newblock


\bibitem[Hamm(2017)]%
        {hamm2017minimax}
\bibfield{author}{\bibinfo{person}{Jihun Hamm}.}
  \bibinfo{year}{2017}\natexlab{}.
\newblock \showarticletitle{Minimax filter: Learning to preserve privacy from
  inference attacks}.
\newblock \bibinfo{journal}{\emph{The Journal of Machine Learning Research}}
  \bibinfo{volume}{18}, \bibinfo{number}{1} (\bibinfo{year}{2017}),
  \bibinfo{pages}{4704--4734}.
\newblock


\bibitem[He et~al\mbox{.}(2015)]%
        {he2015dpt}
\bibfield{author}{\bibinfo{person}{Xi He}, \bibinfo{person}{Graham Cormode},
  \bibinfo{person}{Ashwin Machanavajjhala}, \bibinfo{person}{Cecilia~M
  Procopiuc}, {and} \bibinfo{person}{Divesh Srivastava}.}
  \bibinfo{year}{2015}\natexlab{}.
\newblock \showarticletitle{DPT: differentially private trajectory synthesis
  using hierarchical reference systems}.
\newblock \bibinfo{journal}{\emph{Proceedings of the VLDB Endowment}}
  \bibinfo{volume}{8}, \bibinfo{number}{11} (\bibinfo{year}{2015}),
  \bibinfo{pages}{1154--1165}.
\newblock


\bibitem[Hochreiter and Schmidhuber(1997)]%
        {hochreiter1997long}
\bibfield{author}{\bibinfo{person}{Sepp Hochreiter} {and}
  \bibinfo{person}{J{\"u}rgen Schmidhuber}.} \bibinfo{year}{1997}\natexlab{}.
\newblock \showarticletitle{Long short-term memory}.
\newblock \bibinfo{journal}{\emph{Neural computation}} \bibinfo{volume}{9},
  \bibinfo{number}{8} (\bibinfo{year}{1997}), \bibinfo{pages}{1735--1780}.
\newblock


\bibitem[Huang et~al\mbox{.}(2019)]%
        {huang_variational_2019}
\bibfield{author}{\bibinfo{person}{Dou Huang}, \bibinfo{person}{Xuan Song},
  \bibinfo{person}{Zipei Fan}, \bibinfo{person}{Renhe Jiang},
  \bibinfo{person}{Ryosuke Shibasaki}, \bibinfo{person}{Yu Zhang},
  \bibinfo{person}{Haizhong Wang}, {and} \bibinfo{person}{Yugo Kato}.}
  \bibinfo{year}{2019}\natexlab{}.
\newblock \showarticletitle{A variational autoencoder based generative model of
  urban human mobility}. In \bibinfo{booktitle}{\emph{2019 IEEE Conference on
  Multimedia Information Processing and Retrieval (MIPR)}}.
  \bibinfo{publisher}{{IEEE}}, \bibinfo{address}{San Jose, CA, USA},
  \bibinfo{pages}{425--430}.
\newblock


\bibitem[Huang et~al\mbox{.}(2018)]%
        {huang2018location}
\bibfield{author}{\bibinfo{person}{Haosheng Huang}, \bibinfo{person}{Georg
  Gartner}, \bibinfo{person}{Jukka~M Krisp}, \bibinfo{person}{Martin Raubal},
  {and} \bibinfo{person}{Nico Van~de Weghe}.} \bibinfo{year}{2018}\natexlab{}.
\newblock \showarticletitle{Location based services: ongoing evolution and
  research agenda}.
\newblock \bibinfo{journal}{\emph{Journal of Location Based Services}}
  \bibinfo{volume}{12}, \bibinfo{number}{2} (\bibinfo{year}{2018}),
  \bibinfo{pages}{63--93}.
\newblock


\bibitem[Jiang et~al\mbox{.}(2021)]%
        {jiang2021location}
\bibfield{author}{\bibinfo{person}{Hongbo Jiang}, \bibinfo{person}{Jie Li},
  \bibinfo{person}{Ping Zhao}, \bibinfo{person}{Fanzi Zeng},
  \bibinfo{person}{Zhu Xiao}, {and} \bibinfo{person}{Arun Iyengar}.}
  \bibinfo{year}{2021}\natexlab{}.
\newblock \showarticletitle{Location privacy-preserving mechanisms in
  location-based services: A comprehensive survey}.
\newblock \bibinfo{journal}{\emph{ACM Computing Surveys (CSUR)}}
  \bibinfo{volume}{54}, \bibinfo{number}{1} (\bibinfo{year}{2021}),
  \bibinfo{pages}{1--36}.
\newblock


\bibitem[Kolodziej and Hjelm(2017)]%
        {kolodziej2017local}
\bibfield{author}{\bibinfo{person}{Krzysztof~W Kolodziej} {and}
  \bibinfo{person}{Johan Hjelm}.} \bibinfo{year}{2017}\natexlab{}.
\newblock \bibinfo{booktitle}{\emph{Local positioning systems: LBS applications
  and services}}.
\newblock \bibinfo{publisher}{CRC press}, \bibinfo{address}{Boca Raton}.
\newblock


\bibitem[Krumm(2009)]%
        {krumm2009survey}
\bibfield{author}{\bibinfo{person}{John Krumm}.}
  \bibinfo{year}{2009}\natexlab{}.
\newblock \showarticletitle{A survey of computational location privacy}.
\newblock \bibinfo{journal}{\emph{Personal and Ubiquitous Computing}}
  \bibinfo{volume}{13}, \bibinfo{number}{6} (\bibinfo{year}{2009}),
  \bibinfo{pages}{391--399}.
\newblock


\bibitem[Laurila et~al\mbox{.}(2012)]%
        {laurila2012mobile}
\bibfield{author}{\bibinfo{person}{J.~K. Laurila}, \bibinfo{person}{Daniel
  Gatica-Perez}, \bibinfo{person}{I. Aad}, \bibinfo{person}{Blom J.},
  \bibinfo{person}{Olivier Bornet}, \bibinfo{person}{Trinh-Minh-Tri Do},
  \bibinfo{person}{O. Dousse}, \bibinfo{person}{J. Eberle}, {and}
  \bibinfo{person}{M. Miettinen}.} \bibinfo{year}{2012}\natexlab{}.
\newblock \showarticletitle{The Mobile Data Challenge: Big Data for Mobile
  Computing Research}. In \bibinfo{booktitle}{\emph{Proceedings of the
  Conjunction with the 10th International Conference on Pervasive Computing}}.
  \bibinfo{publisher}{EPFL}, \bibinfo{address}{Newcastle, UK},
  \bibinfo{pages}{1–8}.
\newblock
\urldef\tempurl%
\url{http://infoscience.epfl.ch/record/192489}
\showURL{%
\tempurl}


\bibitem[Lippmann et~al\mbox{.}(2000)]%
        {lippmann2000evaluating}
\bibfield{author}{\bibinfo{person}{Richard~P Lippmann},
  \bibinfo{person}{David~J Fried}, \bibinfo{person}{Isaac Graf},
  \bibinfo{person}{Joshua~W Haines}, \bibinfo{person}{Kristopher~R Kendall},
  \bibinfo{person}{David McClung}, \bibinfo{person}{Dan Weber},
  \bibinfo{person}{Seth~E Webster}, \bibinfo{person}{Dan Wyschogrod},
  \bibinfo{person}{Robert~K Cunningham}, {et~al\mbox{.}}}
  \bibinfo{year}{2000}\natexlab{}.
\newblock \showarticletitle{Evaluating intrusion detection systems: The 1998
  DARPA off-line intrusion detection evaluation}. In
  \bibinfo{booktitle}{\emph{Proceedings DARPA Information Survivability
  Conference and Exposition. DISCEX'00}}, Vol.~\bibinfo{volume}{2}. IEEE,
  \bibinfo{pages}{12--26}.
\newblock


\bibitem[Liu et~al\mbox{.}(2019)]%
        {liu2019privacy}
\bibfield{author}{\bibinfo{person}{Sicong Liu}, \bibinfo{person}{Junzhao Du},
  \bibinfo{person}{Anshumali Shrivastava}, {and} \bibinfo{person}{Lin Zhong}.}
  \bibinfo{year}{2019}\natexlab{}.
\newblock \showarticletitle{Privacy adversarial network: representation
  learning for mobile data privacy}.
\newblock \bibinfo{journal}{\emph{Proceedings of the ACM on Interactive,
  Mobile, Wearable and Ubiquitous Technologies}} \bibinfo{volume}{3},
  \bibinfo{number}{4} (\bibinfo{year}{2019}), \bibinfo{pages}{1--18}.
\newblock


\bibitem[Luca et~al\mbox{.}(2021)]%
        {luca2021survey}
\bibfield{author}{\bibinfo{person}{Massimiliano Luca}, \bibinfo{person}{Gianni
  Barlacchi}, \bibinfo{person}{Bruno Lepri}, {and} \bibinfo{person}{Luca
  Pappalardo}.} \bibinfo{year}{2021}\natexlab{}.
\newblock \bibinfo{title}{A Survey on Deep Learning for Human Mobility}.
\newblock
\newblock
\showeprint[arxiv]{2012.02825}~[cs.LG]


\bibitem[Malekzadeh et~al\mbox{.}(2020)]%
        {malekzadeh2020privacy}
\bibfield{author}{\bibinfo{person}{Mohammad Malekzadeh},
  \bibinfo{person}{Richard~G Clegg}, \bibinfo{person}{Andrea Cavallaro}, {and}
  \bibinfo{person}{Hamed Haddadi}.} \bibinfo{year}{2020}\natexlab{}.
\newblock \showarticletitle{Privacy and utility preserving sensor-data
  transformations}.
\newblock \bibinfo{journal}{\emph{Pervasive and Mobile Computing}}
  \bibinfo{volume}{63} (\bibinfo{year}{2020}), \bibinfo{pages}{101132}.
\newblock


\bibitem[Mukherjee et~al\mbox{.}(2021)]%
        {mukherjee2021privgan}
\bibfield{author}{\bibinfo{person}{Sumit Mukherjee}, \bibinfo{person}{Yixi Xu},
  \bibinfo{person}{Anusua Trivedi}, \bibinfo{person}{Nabajyoti Patowary}, {and}
  \bibinfo{person}{Juan~L Ferres}.} \bibinfo{year}{2021}\natexlab{}.
\newblock \showarticletitle{privGAN: Protecting GANs from membership inference
  attacks at low cost to utility}.
\newblock \bibinfo{journal}{\emph{Proceedings on Privacy Enhancing
  Technologies}} \bibinfo{volume}{2021}, \bibinfo{number}{3}
  (\bibinfo{year}{2021}), \bibinfo{pages}{142--163}.
\newblock


\bibitem[Nasr et~al\mbox{.}(2018)]%
        {nasr2018machine}
\bibfield{author}{\bibinfo{person}{Milad Nasr}, \bibinfo{person}{Reza Shokri},
  {and} \bibinfo{person}{Amir Houmansadr}.} \bibinfo{year}{2018}\natexlab{}.
\newblock \showarticletitle{Machine Learning with Membership Privacy Using
  Adversarial Regularization}. In \bibinfo{booktitle}{\emph{Proceedings of the
  2018 ACM SIGSAC Conference on Computer and Communications Security}}
  (Toronto, Canada) \emph{(\bibinfo{series}{CCS '18})}.
  \bibinfo{publisher}{Association for Computing Machinery},
  \bibinfo{address}{New York, NY, USA}, \bibinfo{pages}{634–646}.
\newblock
\showISBNx{9781450356930}
\urldef\tempurl%
\url{https://doi.org/10.1145/3243734.3243855}
\showDOI{\tempurl}


\bibitem[Oliver et~al\mbox{.}(2020)]%
        {oliver2020mobile}
\bibfield{author}{\bibinfo{person}{Nuria Oliver}, \bibinfo{person}{Bruno
  Lepri}, \bibinfo{person}{Harald Sterly}, \bibinfo{person}{Renaud Lambiotte},
  \bibinfo{person}{S{\'e}bastien Deletaille}, \bibinfo{person}{Marco De~Nadai},
  \bibinfo{person}{Emmanuel Letouz{\'e}}, \bibinfo{person}{Albert~Ali Salah},
  \bibinfo{person}{Richard Benjamins}, \bibinfo{person}{Ciro Cattuto},
  {et~al\mbox{.}}} \bibinfo{year}{2020}\natexlab{}.
\newblock \bibinfo{title}{Mobile phone data for informing public health actions
  across the COVID-19 pandemic life cycle}.
\newblock
\newblock


\bibitem[Oliver et~al\mbox{.}(2015)]%
        {oliver2015mobile}
\bibfield{author}{\bibinfo{person}{Nuria Oliver}, \bibinfo{person}{Aleksandar
  Matic}, {and} \bibinfo{person}{Enrique Frias-Martinez}.}
  \bibinfo{year}{2015}\natexlab{}.
\newblock \showarticletitle{Mobile network data for public health:
  opportunities and challenges}.
\newblock \bibinfo{journal}{\emph{Frontiers in public health}}
  \bibinfo{volume}{3} (\bibinfo{year}{2015}), \bibinfo{pages}{189}.
\newblock


\bibitem[Ouyang et~al\mbox{.}(2018)]%
        {ijcai2018-530}
\bibfield{author}{\bibinfo{person}{Kun Ouyang}, \bibinfo{person}{Reza Shokri},
  \bibinfo{person}{David~S. Rosenblum}, {and} \bibinfo{person}{Wenzhuo Yang}.}
  \bibinfo{year}{2018}\natexlab{}.
\newblock \showarticletitle{A Non-Parametric Generative Model for Human
  Trajectories}. In \bibinfo{booktitle}{\emph{Proceedings of the Twenty-Seventh
  International Joint Conference on Artificial Intelligence, {IJCAI-18}}}.
  \bibinfo{publisher}{International Joint Conferences on Artificial
  Intelligence Organization}, \bibinfo{address}{Stockholm, Sweden},
  \bibinfo{pages}{3812--3817}.
\newblock
\urldef\tempurl%
\url{https://doi.org/10.24963/ijcai.2018/530}
\showDOI{\tempurl}


\bibitem[Primault et~al\mbox{.}(2018)]%
        {primault2018long}
\bibfield{author}{\bibinfo{person}{Vincent Primault}, \bibinfo{person}{Antoine
  Boutet}, \bibinfo{person}{Sonia~Ben Mokhtar}, {and} \bibinfo{person}{Lionel
  Brunie}.} \bibinfo{year}{2018}\natexlab{}.
\newblock \showarticletitle{The long road to computational location privacy: A
  survey}.
\newblock \bibinfo{journal}{\emph{IEEE Communications Surveys \& Tutorials}}
  \bibinfo{volume}{21}, \bibinfo{number}{3} (\bibinfo{year}{2018}),
  \bibinfo{pages}{2772--2793}.
\newblock


\bibitem[Primault et~al\mbox{.}(2014)]%
        {primault2014differentially}
\bibfield{author}{\bibinfo{person}{Vincent Primault},
  \bibinfo{person}{Sonia~Ben Mokhtar}, \bibinfo{person}{C{\'e}dric Lauradoux},
  {and} \bibinfo{person}{Lionel Brunie}.} \bibinfo{year}{2014}\natexlab{}.
\newblock \bibinfo{title}{Differentially Private Location Privacy in Practice}.
\newblock
\newblock
\urldef\tempurl%
\url{https://doi.org/10.48550/ARXIV.1410.7744}
\showDOI{\tempurl}


\bibitem[Rao et~al\mbox{.}(2020)]%
        {rao2020lstm}
\bibfield{author}{\bibinfo{person}{Jinmeng Rao}, \bibinfo{person}{Song Gao},
  \bibinfo{person}{Yuhao Kang}, {and} \bibinfo{person}{Qunying Huang}.}
  \bibinfo{year}{2020}\natexlab{}.
\newblock \bibinfo{title}{LSTM-TrajGAN: A Deep Learning Approach to Trajectory
  Privacy Protection}.
\newblock
\newblock
\urldef\tempurl%
\url{https://doi.org/10.48550/ARXIV.2006.10521}
\showDOI{\tempurl}


\bibitem[Rezaei et~al\mbox{.}(2018)]%
        {rezaei2018protecting}
\bibfield{author}{\bibinfo{person}{Aria Rezaei}, \bibinfo{person}{Chaowei
  Xiao}, \bibinfo{person}{Jie Gao}, \bibinfo{person}{Bo Li}, {and}
  \bibinfo{person}{Sirajum Munir}.} \bibinfo{year}{2018}\natexlab{}.
\newblock \bibinfo{title}{Application-driven Privacy-preserving Data Publishing
  with Correlated Attributes}.
\newblock
\newblock
\urldef\tempurl%
\url{https://doi.org/10.48550/ARXIV.1812.10193}
\showDOI{\tempurl}


\bibitem[Shin et~al\mbox{.}(2020)]%
        {shin2020user}
\bibfield{author}{\bibinfo{person}{Seungjae Shin}, \bibinfo{person}{Hongseok
  Jeon}, \bibinfo{person}{Chunglae Cho}, \bibinfo{person}{Seunghyun Yoon},
  {and} \bibinfo{person}{Taeyeon Kim}.} \bibinfo{year}{2020}\natexlab{}.
\newblock \showarticletitle{User Mobility Synthesis based on Generative
  Adversarial Networks: A Survey}. In \bibinfo{booktitle}{\emph{2020 22nd
  International Conference on Advanced Communication Technology (ICACT)}}.
  \bibinfo{publisher}{{IEEE}}, \bibinfo{address}{Phoenix Park, Korea (South)},
  \bibinfo{pages}{94--103}.
\newblock


\bibitem[Shokri et~al\mbox{.}(2011)]%
        {shokri2011quantifying}
\bibfield{author}{\bibinfo{person}{Reza Shokri}, \bibinfo{person}{George
  Theodorakopoulos}, \bibinfo{person}{Jean-Yves Le~Boudec}, {and}
  \bibinfo{person}{Jean-Pierre Hubaux}.} \bibinfo{year}{2011}\natexlab{}.
\newblock \showarticletitle{Quantifying location privacy}. In
  \bibinfo{booktitle}{\emph{2011 IEEE symposium on security and privacy}}.
  \bibinfo{publisher}{{IEEE}}, \bibinfo{address}{Oakland, CA, USA},
  \bibinfo{pages}{247--262}.
\newblock


\bibitem[Shokri et~al\mbox{.}(2012)]%
        {shokri2012protecting}
\bibfield{author}{\bibinfo{person}{Reza Shokri}, \bibinfo{person}{George
  Theodorakopoulos}, \bibinfo{person}{Carmela Troncoso},
  \bibinfo{person}{Jean-Pierre Hubaux}, {and} \bibinfo{person}{Jean-Yves
  Le~Boudec}.} \bibinfo{year}{2012}\natexlab{}.
\newblock \showarticletitle{Protecting Location Privacy: Optimal Strategy
  against Localization Attacks}. In \bibinfo{booktitle}{\emph{Proceedings of
  the 2012 ACM Conference on Computer and Communications Security}} (Raleigh,
  North Carolina, USA) \emph{(\bibinfo{series}{CCS '12})}.
  \bibinfo{publisher}{Association for Computing Machinery},
  \bibinfo{address}{New York, NY, USA}, \bibinfo{pages}{617–627}.
\newblock
\showISBNx{9781450316514}
\urldef\tempurl%
\url{https://doi.org/10.1145/2382196.2382261}
\showDOI{\tempurl}


\bibitem[Song et~al\mbox{.}(2010)]%
        {song2010limits}
\bibfield{author}{\bibinfo{person}{Chaoming Song}, \bibinfo{person}{Zehui Qu},
  \bibinfo{person}{Nicholas Blumm}, {and}
  \bibinfo{person}{Albert-L{\'a}szl{\'o} Barab{\'a}si}.}
  \bibinfo{year}{2010}\natexlab{}.
\newblock \showarticletitle{Limits of predictability in human mobility}.
\newblock \bibinfo{journal}{\emph{Science}} \bibinfo{volume}{327},
  \bibinfo{number}{5968} (\bibinfo{year}{2010}), \bibinfo{pages}{1018--1021}.
\newblock


\bibitem[Toch et~al\mbox{.}(2019)]%
        {toch2019analyzing}
\bibfield{author}{\bibinfo{person}{Eran Toch}, \bibinfo{person}{Boaz Lerner},
  \bibinfo{person}{Eyal Ben-Zion}, {and} \bibinfo{person}{Irad Ben-Gal}.}
  \bibinfo{year}{2019}\natexlab{}.
\newblock \showarticletitle{Analyzing large-scale human mobility data: a survey
  of machine learning methods and applications}.
\newblock \bibinfo{journal}{\emph{Knowledge and Information Systems}}
  \bibinfo{volume}{58}, \bibinfo{number}{3} (\bibinfo{year}{2019}),
  \bibinfo{pages}{501--523}.
\newblock


\bibitem[Wang et~al\mbox{.}(2019)]%
        {wang2019achieving}
\bibfield{author}{\bibinfo{person}{Jinbao Wang}, \bibinfo{person}{Zhipeng Cai},
  {and} \bibinfo{person}{Jiguo Yu}.} \bibinfo{year}{2019}\natexlab{}.
\newblock \showarticletitle{Achieving personalized $ k $-anonymity-based
  content privacy for autonomous vehicles in CPS}.
\newblock \bibinfo{journal}{\emph{IEEE Transactions on Industrial Informatics}}
  \bibinfo{volume}{16}, \bibinfo{number}{6} (\bibinfo{year}{2019}),
  \bibinfo{pages}{4242--4251}.
\newblock


\bibitem[Wang et~al\mbox{.}(2020)]%
        {wang2020deep}
\bibfield{author}{\bibinfo{person}{Senzhang Wang}, \bibinfo{person}{Jiannong
  Cao}, {and} \bibinfo{person}{Philip Yu}.} \bibinfo{year}{2020}\natexlab{}.
\newblock \showarticletitle{Deep Learning for Spatio-Temporal Data Mining: A
  Survey}.
\newblock \bibinfo{journal}{\emph{IEEE Transactions on Knowledge and Data
  Engineering}} \bibinfo{volume}{14}, \bibinfo{number}{8}
  (\bibinfo{year}{2020}), \bibinfo{pages}{1--1}.
\newblock
\urldef\tempurl%
\url{https://doi.org/10.1109/TKDE.2020.3025580}
\showDOI{\tempurl}


\bibitem[Xiao and Xiong(2015)]%
        {xiao2015protecting}
\bibfield{author}{\bibinfo{person}{Yonghui Xiao} {and} \bibinfo{person}{Li
  Xiong}.} \bibinfo{year}{2015}\natexlab{}.
\newblock \showarticletitle{Protecting Locations with Differential Privacy
  under Temporal Correlations}. In \bibinfo{booktitle}{\emph{Proceedings of the
  22nd ACM SIGSAC Conference on Computer and Communications Security}} (Denver,
  Colorado, USA) \emph{(\bibinfo{series}{CCS '15})}.
  \bibinfo{publisher}{Association for Computing Machinery},
  \bibinfo{address}{New York, NY, USA}, \bibinfo{pages}{1298–1309}.
\newblock
\showISBNx{9781450338325}
\urldef\tempurl%
\url{https://doi.org/10.1145/2810103.2813640}
\showDOI{\tempurl}


\bibitem[Xiao et~al\mbox{.}(2017)]%
        {xiao_loclok_2017}
\bibfield{author}{\bibinfo{person}{Yonghui Xiao}, \bibinfo{person}{Li Xiong},
  \bibinfo{person}{Si Zhang}, {and} \bibinfo{person}{Yang Cao}.}
  \bibinfo{year}{2017}\natexlab{}.
\newblock \showarticletitle{LocLok: location cloaking with differential privacy
  via hidden markov model}.
\newblock \bibinfo{journal}{\emph{Proceedings of the VLDB Endowment}}
  \bibinfo{volume}{10} (\bibinfo{date}{08} \bibinfo{year}{2017}),
  \bibinfo{pages}{1901--1904}.
\newblock
\urldef\tempurl%
\url{https://doi.org/10.14778/3137765.3137804}
\showDOI{\tempurl}


\bibitem[Yang et~al\mbox{.}(2014)]%
        {dingqi_yang_modeling_2015}
\bibfield{author}{\bibinfo{person}{Dingqi Yang}, \bibinfo{person}{Daqing
  Zhang}, \bibinfo{person}{Vincent~W Zheng}, {and} \bibinfo{person}{Zhiyong
  Yu}.} \bibinfo{year}{2014}\natexlab{}.
\newblock \showarticletitle{Modeling user activity preference by leveraging
  user spatial temporal characteristics in {LBSNs}}.
\newblock \bibinfo{journal}{\emph{IEEE Transactions on Systems, Man, and
  Cybernetics: Systems}} \bibinfo{volume}{45}, \bibinfo{number}{1}
  (\bibinfo{year}{2014}), \bibinfo{pages}{129--142}.
\newblock


\bibitem[Yang et~al\mbox{.}(2021)]%
        {yang_pptpf_2021}
\bibfield{author}{\bibinfo{person}{Jianxi Yang}, \bibinfo{person}{Manoranjan
  Dash}, {and} \bibinfo{person}{Sin~G. Teo}.} \bibinfo{year}{2021}\natexlab{}.
\newblock \showarticletitle{{PPTPF}: {Privacy}-{Preserving} {Trajectory}
  {Publication} {Framework} for {CDR} {Mobile} {Trajectories}}.
\newblock \bibinfo{journal}{\emph{ISPRS International Journal of
  Geo-Information}} \bibinfo{volume}{10}, \bibinfo{number}{4}
  (\bibinfo{date}{April} \bibinfo{year}{2021}), \bibinfo{pages}{224}.
\newblock
\urldef\tempurl%
\url{https://doi.org/10.3390/ijgi10040224}
\showDOI{\tempurl}
\newblock
\shownote{Number: 4 Publisher: Multidisciplinary Digital Publishing Institute}.


\bibitem[Zhan and Haddadi(2019)]%
        {zhan2019towards}
\bibfield{author}{\bibinfo{person}{Yuting Zhan} {and} \bibinfo{person}{Hamed
  Haddadi}.} \bibinfo{year}{2019}\natexlab{}.
\newblock \showarticletitle{Towards Automating Smart Homes: Contextual and
  Temporal Dynamics of Activity Prediction}. In
  \bibinfo{booktitle}{\emph{Adjunct Proceedings of the 2019 ACM International
  Joint Conference on Pervasive and Ubiquitous Computing and Proceedings of the
  2019 ACM International Symposium on Wearable Computers}} (London, United
  Kingdom) \emph{(\bibinfo{series}{UbiComp/ISWC '19 Adjunct})}.
  \bibinfo{publisher}{Association for Computing Machinery},
  \bibinfo{address}{New York, NY, USA}, \bibinfo{pages}{413–417}.
\newblock
\showISBNx{9781450368698}
\urldef\tempurl%
\url{https://doi.org/10.1145/3341162.3349307}
\showDOI{\tempurl}


\bibitem[Zheng et~al\mbox{.}(2011)]%
        {zheng2011geolife}
\bibfield{author}{\bibinfo{person}{Yu Zheng}, \bibinfo{person}{Hao Fu},
  \bibinfo{person}{Xing Xie}, \bibinfo{person}{Wei-Ying Ma}, {and}
  \bibinfo{person}{Quannan Li}.} \bibinfo{year}{2011}\natexlab{}.
\newblock \bibinfo{booktitle}{\emph{Geolife GPS trajectory dataset - User
  Guide} (\bibinfo{edition}{geolife gps trajectories 1.1} ed.)}.
\newblock Microsoft Research Asia.
\newblock
\urldef\tempurl%
\url{https://www.microsoft.com/en-us/research/publication/geolife-gps-trajectory-dataset-user-guide/}
\showURL{%
\tempurl}
\newblock
\shownote{Geolife GPS trajectories 1.1}.


\end{thebibliography}

\end{document}